\documentclass[11pt]{article}
\PassOptionsToPackage{table,dvipsnames}{xcolor}
\usepackage{smureport}

\usepackage{amsmath,amsfonts,bm}

\def\eqref#1{equation~\ref{#1}}

\def\1{\bm{1}}

\DeclareMathAlphabet{\mathsfit}{\encodingdefault}{\sfdefault}{m}{sl}
\SetMathAlphabet{\mathsfit}{bold}{\encodingdefault}{\sfdefault}{bx}{n}

\usepackage[utf8]{inputenc} 
\usepackage[T1]{fontenc}    
\usepackage{hyperref}       
\usepackage{xurl}            
\usepackage{booktabs}       
\usepackage{amsfonts}       
\usepackage{nicefrac}       
\usepackage{microtype}      

\usepackage{graphicx}
\usepackage{tcolorbox}
\tcbuselibrary{breakable}
\usepackage{multirow}
\usepackage{amsmath}
\usepackage{xcolor}
\usepackage{collcell}
\usepackage{pgf}
\usepackage{algorithm}
\usepackage{algorithmic}
\usepackage{caption}
\usepackage{enumitem}

\usepackage{wrapfig}

\usepackage{makecell}

\newcommand{\pos}[1]{\textcolor{ForestGreen}{\scriptsize (+#1)}}

\title{Video-Browser: Towards Open-web Agentic Video Browsing}

\definecolor{highacc}{HTML}{2e8b57}

\newcommand{\heat}[1]{%
    \pgfmathparse{#1 > 0.01 ? 1 : 0}%
    \ifnum\pgfmathresult=1
        \pgfmathparse{min(100, #1 * 1.4)}%
        \let\intensity\pgfmathresult
        \edef\x{\noexpand\cellcolor{highacc!\intensity}}\x%
        #1%
    \else
        \cellcolor{white}#1%
    \fi
}

\author{%
  Zhengyang Liang\textsuperscript{1,$\clubsuit$}\quad
  Yan Shu\textsuperscript{2,$\clubsuit$}\quad
  Xiangrui Liu\textsuperscript{3}\quad
  Minghao Qin\textsuperscript{3}\\[2pt]
  Nicu Sebe\textsuperscript{2}\quad
  Jiaxin Mao\textsuperscript{4}\quad
  Zheng Liu\textsuperscript{5}\quad
  Lizi Liao\textsuperscript{1}}
\smubrand{Research Reports}
\smuinstitute{}
\smuaffiliations{%
  \textsuperscript{1}Singapore Management University\quad
  \textsuperscript{2}University of Trento\\[2pt]
  \textsuperscript{3}Beijing Academy of Artificial Intelligence\quad
  \textsuperscript{4}Renmin University of China\\[2pt]
  \textsuperscript{5}Hong Kong Polytechnic University}
\smuauthorsnote{\textsuperscript{$\clubsuit$}Equal contribution}
\smudate{\today}
\smucontact{\href{mailto:zyliang@smu.edu.sg}{zyliang@smu.edu.sg};
  \href{mailto:yan.shu@unitn.it}{yan.shu@unitn.it};
  \href{mailto:lzliao@smu.edu.sg}{lzliao@smu.edu.sg}}
\smunote{\textbf{Code:}
  \url{https://github.com/chrisx599/Video-Browser}\\
  \textbf{Project:}
  \url{https://liang-zhengyang.github.io/video-browsecomp/}\\
  \textbf{Benchmark:}
  \url{https://huggingface.co/datasets/chr1ce/Video-Browsecomp}}
\smushorttitle{Video-Browser: Towards Open-web Agentic Video Browsing}
\smuheaderlogos{%
  \begin{minipage}[c]{30mm}\raggedright
    \includegraphics[width=30mm]{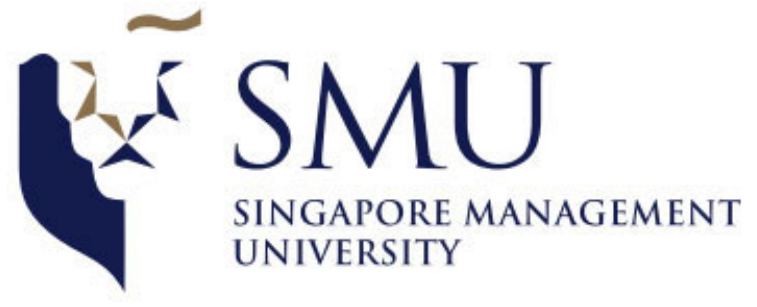}
  \end{minipage}\hfill%
  \begin{minipage}[c]{27mm}\centering
    \includegraphics[width=27mm]{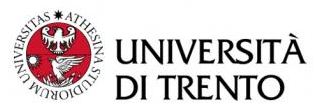}
  \end{minipage}\hfill%
  \begin{minipage}[c]{23mm}\centering
    \includegraphics[width=23mm]{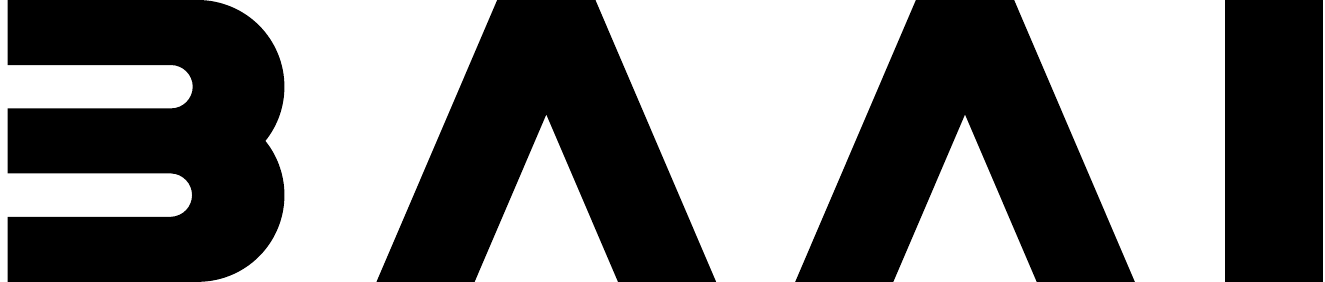}
  \end{minipage}\hfill%
  \begin{minipage}[c]{36mm}\centering
    \includegraphics[width=36mm]{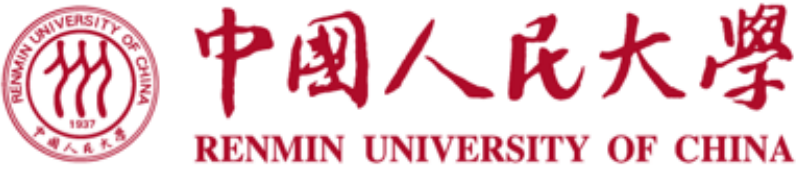}
  \end{minipage}\hfill%
  \begin{minipage}[c]{37mm}\raggedleft
    \includegraphics[width=37mm]{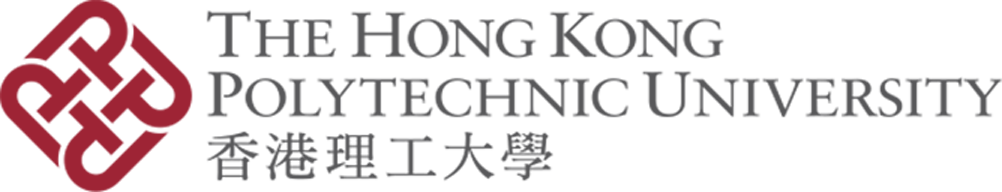}
  \end{minipage}}
\hypersetup{pdftitle={Video-Browser: Towards Open-web Agentic Video Browsing},
  pdfauthor={Zhengyang Liang, Yan Shu, Xiangrui Liu, Minghao Qin,
    Nicu Sebe, Jiaxin Mao, Zheng Liu, Lizi Liao},
  pdfsubject={Video-BrowseComp and Video-Browser}}

\begin{document}

\maketitle

\begin{smuabstract}\small
Autonomous agents are shifting information seeking from passive retrieval toward proactive, open-ended research, yet this progress has largely bypassed video, one of the web's richest sources of evidence. We formalize Agentic Video Browsing, in which an agent must query video search engines, decide which returned videos to watch, and answer questions grounded in their visual content. To study this setting reproducibly, we introduce \textbf{Video-BrowseComp}, a benchmark and infrastructure suite with 660 human-annotated questions at three difficulty levels, paired online and offline evaluation tracks, and a fixed video search stack of roughly 12,000 videos spanning over 2,000 hours for evaluation and training-data construction. Our analysis reveals that the main bottleneck lies before video understanding: agents often fail to find the answer-bearing video at all. This bottleneck stems from two coupled challenges. Search engines index textual metadata such as titles and tags, whereas questions often target implicit visual or cross-source evidence absent from these surface forms; and exhaustively inspecting the many returned videos is prohibitively expensive. We propose \textbf{Video-Browser}, an evidence-surfacing agent that improves both sides of search. A Translation Memory helps the Planner phrase queries by analogy to past question-to-query transformations that successfully surfaced their target videos, and a Watcher performs sparse visual screening over returned candidates, retaining only videos that plausibly contain answer-bearing evidence for later dense verification. On the offline track, Video-Browser raises overall accuracy from 11.6\% to 27.3\% with Qwen3-VL-8B and from 39.8\% to 45.1\% with Gemini-3.1-Pro over the corresponding ReAct search agents, showing that better query formulation and post-search candidate selection improve end-to-end browsing performance.
We open-source all the data and agent code.
\end{smuabstract}
\clearpage

\section{Introduction}
\label{sec:introduction}

The rapid progress of Large Language Models (LLMs) has shifted information seeking from static question answering toward autonomous agents that actively navigate the web to solve complex problems~\citep{nakano2021webgpt, he2024webvoyager}.
Much of this progress has focused on text and images~\citep{jin2025searchr1trainingllmsreason, geng2025webwatcherbreakingnewfrontier}.
Yet a substantial amount of information on the web is conveyed visually rather than through text alone: instructional videos demonstrate how a task is performed, lectures explain concepts visually, and recordings document what happened during an event.
Such evidence often exists only inside the video stream, so an agent that can find, watch, and reason over open-web videos would answer a class of questions that text-centric agents cannot. Video nevertheless remains largely absent from autonomous, open-ended information seeking.

Existing research leaves this setting uncovered from two directions. On the video side, video understanding models~\citep{Shu_2025_CVPR_videoxl} and agents~\citep{wang2024videoagentlongformvideounderstanding} answer questions about videos that have already been provided, while video retrieval systems~\citep{chen2024verifiedvideocorpusmoment, rossetto2025results2025videobrowser} locate videos or moments within a predefined collection indexed in advance.
Both assume that the system already holds the videos, an assumption that is difficult to extend to the open web, where pre-collecting and processing the relevant video space is impractical and the agent must first discover which videos contain the needed information.
On the agent side, WebArena~\citep{zhou2024webarenarealisticwebenvironment} evaluates agents that operate websites for predominantly text-based tasks, and VideoWebArena~\citep{jang2025videowebarena} adds tasks guided by a provided video demonstration, where the video serves as an input to consult rather than evidence to be discovered.
BrowseComp~\citep{wei2025browsecompsimplechallengingbenchmark} evaluates agents that search the web to answer hard questions, primarily over text, and MM-BrowseComp~\citep{li2025mmbrowsecompcomprehensivebenchmarkmultimodal} and follow-ups (Section~\ref{related:infoseeking}) extend it to multimodal evidence.
We follow this paradigm but place video at the center: in Agentic Video Browsing, an agent must query video search engines, decide which returned videos to watch, and ground its answer in their visual content.

To systematically evaluate agents in this setting, we introduce \textbf{Video-BrowseComp}, a benchmark and infrastructure suite containing 660 human-annotated evaluation questions spanning eight genres and three difficulty levels, together with 1,000 automatically constructed training questions.
Because live-web search results and video availability change over time, relying on live-web evaluation alone makes results difficult to reproduce and failures difficult to diagnose.
We therefore construct a fixed offline video search stack containing roughly 12,000 videos and over 2,000 hours of footage, making search results, candidate pools, and final accuracy directly comparable across agents.
With this infrastructure, we evaluate current agents and find that locating answer-bearing videos is a major bottleneck. The agents fail to find the target video for most questions.

This search bottleneck stems from two coupled reasons, both rooted in how videos are exposed on the web.
First, there is a mismatch between what users ask and what video search engines expose.
User questions often refer to visual content inside videos, such as a particular on-screen detail, an action that unfolds across several clips, or a fact assembled from multiple videos. However video search operates primarily over textual metadata such as titles, descriptions, and tags~\citep{gong2025cardrewriterleveragingknowledgecards}.
The agent must therefore translate implicit visual evidence needs into search engine friendly queries, a mapping that is difficult to infer from the question alone.
Second, unlike a web page or an image, video candidates are expensive to inspect in detail.
Search engines may return many long and visually redundant videos, and densely processing every candidate quickly exceeds practical compute and context budgets.
An effective video browsing agent must therefore not only formulate better search queries, but also determine which returned videos are worth examining further.

To address these bottlenecks, we propose \textbf{Video-Browser}, an evidence-surfacing agent that improves both pre-search query formulation and post-search candidate selection.
Its central design principle is to draw on past successful searches for query formulation and to visually screen returned videos before inspecting them in detail.
Before search, \emph{Translation Memory} provides past search examples that the \emph{Planner} can draw on to phrase new queries by analogy.
These examples are constructed from training questions by generating candidate queries and verifying which ones can surface their corresponding target videos.
After search, a lightweight \emph{Watcher} screens returned candidates using textual metadata and sparse frame grids, retaining videos that plausibly contain the required evidence.
If the accumulated evidence is insufficient, the system identifies what is still missing and feeds this information back to the Planner to guide the next search.
When browsing terminates, the selected videos undergo denser visual verification to produce the final answer.
On the offline track, Video-Browser improves overall accuracy from 11.6\% to 27.3\% with Qwen3-VL-8B and from 39.8\% to 45.1\% with Gemini-3.1-Pro over the corresponding ReAct search agents.

In summary, we make three contributions.

(1). We formalize \textit{Agentic Video Browsing} as an iterative, open-web video-grounded information-seeking task and introduce \textbf{Video-BrowseComp}, a reproducible benchmark and infrastructure suite with questions that require watching videos to answer, three difficulty levels, paired online/offline evaluation tracks, and a large fixed video search stack that also supports training-data construction.

(2). Through a search/reasoning decomposition, we identify video discovery as a major bottleneck before final visual reasoning. Current agents struggle both to translate implicit visual evidence needs into effective search queries and to select evidence-bearing videos from numerous returned candidates without exhaustively inspecting them.

(3). We propose \textbf{Video-Browser}, an evidence-surfacing agent that combines \emph{Translation Memory} for pre-search query formulation with a \emph{Watcher} for post-search visual candidate selection. On the offline track, it improves overall accuracy over the corresponding ReAct search agent from 11.6\% to 27.3\% with Qwen3-VL-8B and from 39.8\% to 45.1\% with Gemini-3.1-Pro.

\section{Related Works}

\subsection{Video Understanding and Retrieval}

The field of video understanding has witnessed rapid advances with the emergence of Multimodal Large Language Models (MLLMs) \citep{comanici2025gemini25pushingfrontier, liu2023visualinstructiontuning, zhu2025internvl3exploringadvancedtraining}. Video LLMs \citep{Shu_2025_CVPR_videoxl, qin2025videoxl2longvideounderstandingtaskaware, li2025videochatflashhierarchicalcompressionlongcontext, Liu_2025_CVPR_nvila, chen2024longvilascalinglongcontextvisual, shen2024longvu} and agents \citep{wang2024videoagentlongformvideounderstanding, fan2024videoagentmemoryaugmentedmultimodalagent, wang2025videotreeadaptivetreebasedvideo, yuan2025videoexplorerthinkvideosagentic, zhang2025deepvideodiscoveryagentic, ren2025videoragretrievalaugmentedgenerationextreme} have demonstrated impressive capabilities in processing long-context visual inputs. 
Concurrently, benchmarks \citep{fu2024video, Zhou_2025_CVPR, Li_2024_CVPR_mvbench, wang2025lvbenchextremelongvideo} have been established to rigorously evaluate these models on traditional video QA tasks.
A complementary line of research studies video retrieval, where systems locate relevant videos or temporal moments from a predefined collection, ranging from embedding-based text-to-video retrieval~\citep{Bain_2021_ICCV, rossetto2025results2025videobrowser} and video-corpus moment retrieval~\citep{chen2024verifiedvideocorpusmoment} to agentic retrieval systems~\citep{wu2025adaptivemultiagentreasoningtexttovideo, Shah_2026_WACV, duong2024mmmsvr, lee2026videosearchr1iterativevideoretrieval}. These approaches typically assume a fixed and fully available corpus that can be processed or indexed in advance. Instead, our setting treats video search as an open-ended information-seeking process: the agent interacts with an external search engine, examines the returned videos as information sources, and continues searching until it can answer the question with sufficient evidence.

\subsection{Agentic Information Seeking}
\label{related:infoseeking}

The development of LLM agents has enabled models to actively interact with external information environments rather than relying solely on provided context. Some works focus on open-web information seeking, where agents iteratively search and leverage external evidence to answer difficult questions~\citep{jin2025searchr1trainingllmsreason, li2025webthinker, wu2025webdancer, searcho1, wei2025browsecompsimplechallengingbenchmark, browsecompzh, mialon2023gaiabenchmarkgeneralai, humanlastexam, chen2025browsecompplusfairtransparentevaluation, li2025webweaverstructuringwebscaleevidence}, 
while others study agents interact and operate web environments~\citep{zhou2024webarenarealisticwebenvironment, koh-etal-2024-visualwebarena, jang2025videowebarena}.
More recently, multimodal information seeking has expanded from text and images to richer visual and video-centric evidence~\citep{li2025mmbrowsecompcomprehensivebenchmarkmultimodal, mmsearch-r1, jiang2024mmsearchbenchmarkingpotentiallarge, peng2025mvuevalmultivideounderstandingevaluation, he2025enhancingvideolargelanguage, fu2025seekingupdatinglivevisual, liu2026watchingreasoningsearchingvideo, liu2026thinkingvideosunifyingvideo, zhang2026adavdradaptivetooluse, fang2026videodeepresearchnextgenerationmultimodaldeepresearch, zeng2026visiondeepresearchbenchmarkrethinkingvisual, zhang2026videogaiabenchmarkgeneralai}.
However, open-web video search introduces a new challenge: search engines primarily expose videos through textual metadata such as titles and tags~\citep{gong2025cardrewriterleveragingknowledgecards}, while the evidence required to answer often lies inside the underlying visual stream. 
Thus, textual relevance does not necessarily imply visual evidence relevance, requiring agents to bridge the gap between search-visible metadata and answer-bearing video content.

\section{Benchmark}
\label{sec:benckmark}

\subsection{Task Definition}

We formulate agentic video browsing as an information-seeking task in which the answer must be grounded in video evidence.
Given a natural language query $Q$, an agent has access to the video search environment $\mathcal{E}$ and must autonomously plan a sequence of actions $S = \{a_1, a_2, ..., a_t\}$, such as searching the web, watching specific video timestamps, or reading subtitles, to generate a final result.
Critically, the agent must both surface answer-bearing videos through effective query formulation and verify that the returned candidates actually contain the required evidence. We further stratify questions into three difficulty levels (Section~\ref{sec:difficulty}) that progressively stress these capabilities, from simple query phrasing to multi-hop reasoning chains.

\begin{figure}[H]
  \centering
   \includegraphics[width=1.0\linewidth]{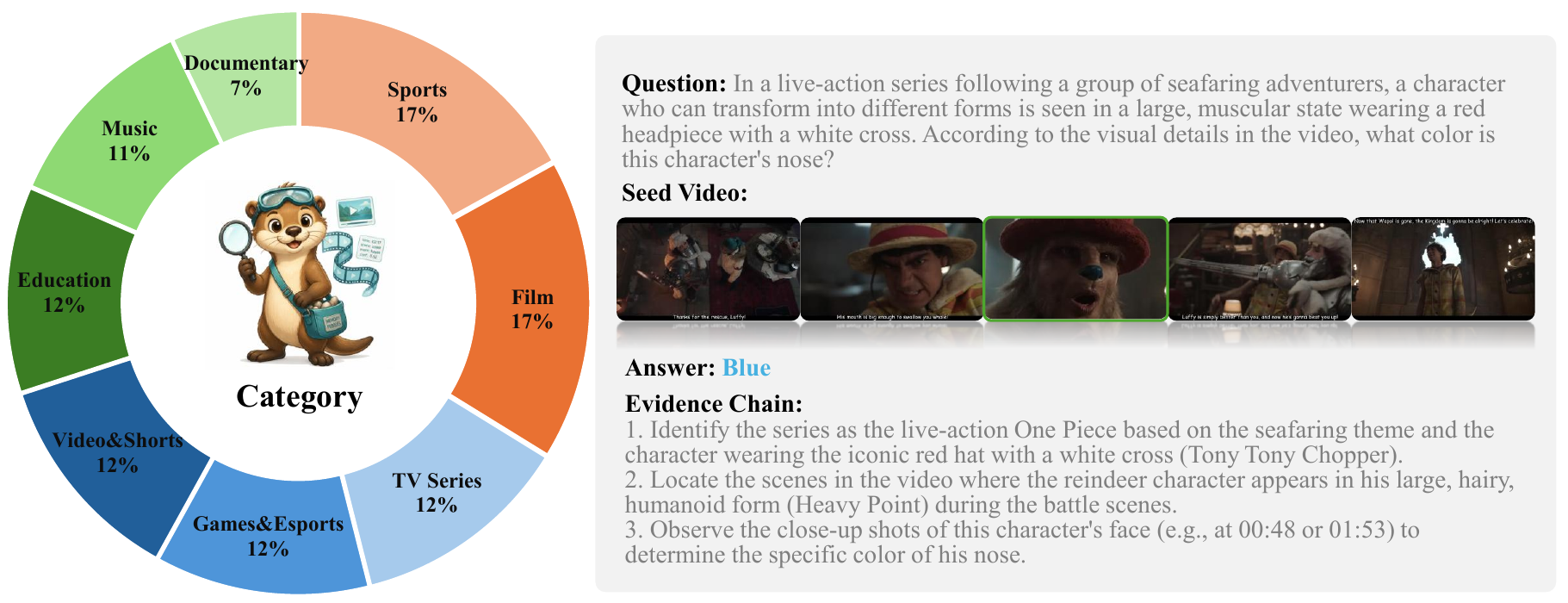}
   
   \caption{
   Statistics and example from Video-BrowseComp. Left: category distribution of evaluation set. Right: an example showing the question, seed video frames, answer, and evidence chain.}
   
   \label{fig:benchmark}
\end{figure}

\subsection{Difficulty Stratification}

\label{sec:difficulty}

To systematically assess the limits of agentic capabilities, we stratify Video-BrowseComp into three distinct difficulty levels based on the complexity of the search space and the depth of reasoning required. We provide detailed benchmark cases in Appendix \ref{sec:bench_samples}.

\textbf{Level 1: Explicit Search.}
Questions in this level provide explicit constraints regarding the search domain, such as specific publication dates, platform names, or title keywords. The challenge lies in instruction following to locate the target video and temporal grounding to extract visual details.\\
\textbf{Level 2: Implicit Search.}
Removing explicit search keywords, the target video is referenced only through indirect descriptions. Agents must first bridge the gap between visual cues and textual search queries, and filter multiple candidate videos to find the one matching the event description.\\
\textbf{Level 3: Multi-Source Reasoning.}
Representing the highest tier of difficulty, these questions cannot be answered by viewing a single video. They require cross-source reasoning where information searched from one source serves as a prerequisite for locating the next.

\subsection{Dataset Construction}

\label{sec:data_pipeline}

To ensure that Video-BrowseComp evaluates agentic video browsing rather than general web search, each question must require visual evidence from video, be hard to find but easy to verify, and have a unique objective answer grounded in observable spatio-temporal evidence. 
The 660-question evaluation set is constructed through a human reverse-annotation pipeline: annotators first identify videos containing verifiable evidence, formulate indirect and compositional questions, and each sample is independently cross-validated by a second annotator. Separately, the 1,000-question training split used to construct Translation Memory is generated through an autonomous reverse-construction pipeline.
For reproducible evaluation, we build an offline search stack in which each offline question is paired with fixed seed videos and cached candidate metadata, including titles, descriptions, channel names, and tags. 
In the offline track, all agents interact with this fixed stack rather than the live web, making search results, candidate pools, and final accuracy directly comparable. 
Full annotation and search-stack construction details are provided in Appendix~\ref{data_pipeline}.

\subsection{Dataset Statistics}

In Figure \ref{fig:benchmark}, Video-BrowseComp consists of 1660 questions spanning 8 diverse genres, partitioned into a 660-question human-annotated evaluation set and a 1000-question automatically constructed training set. To support both open-web use and reproducible academic research, the evaluation set is split into 210 \emph{online} questions for quick evaluation in live-web scenarios and 450 \emph{offline} questions evaluated against our fixed video search stack. This stack contains over 12,000 source videos totaling more than 2,000 hours of footage, and provides cached candidate metadata for the offline track. The 1,000 training questions are also grounded in this stack, with seed videos and searchable metadata drawn from the same fixed corpus. Detailed statistics are provided in Appendix~\ref{detailed_analysis}.

\subsection{Bottleneck Analysis}

\label{sec:bottleneck_analysis}

To pinpoint where the difficulty of Video-BrowseComp originates, we decompose end-to-end accuracy into two separable axes: (1). Search, can the system locate the answer-bearing video at all; (2). Reasoning, given the correct video, can it read the answer off the visual stream. We evaluate these two axes on offline search stack, all backbone index only titles, descriptions, channel names, and tags.

\noindent\textbf{Search axis.}\par\nopagebreak
\noindent Table~\ref{tab:retrieval_decomp} isolates the search bottleneck by varying both the search backbone and the query formulation strategy.

\begin{table}[H]
    \centering
    \caption{Search recall across three backbones and four query forms. \textbf{Bold} marks best performance.}
    \footnotesize
    \setlength{\tabcolsep}{4pt}
    \begin{tabular}{ll cccccc}
        \toprule
        \textbf{Search Method} & \textbf{Query form} & \textbf{R@1} & \textbf{R@5} & \textbf{R@10} & \textbf{R@20} & \textbf{R@50} & \textbf{R@100} \\
        \midrule
        BM25         & raw question                &  3.1 &  6.0 &  7.2 &  9.9 & 11.8 & 15.0 \\
        BM25         & + Qwen3-VL-8B       &  6.9 & 13.5 & 15.5 & 20.4 & 25.7 & 29.6 \\
        BM25         & + GPT-5.4           & 10.7 & 20.7 & 23.9 & 28.0 & 35.7 & 42.4 \\
        BM25         & + Gemini-3.1-Pro    & 15.0 & 24.5 & 29.9 & 33.5 & 39.6 & 44.9 \\
        \midrule
        BGE-M3       & raw question        &  9.3 & 18.8 & 23.2 & 29.0 & 39.2 & 49.0 \\
        BGE-M3       & + Qwen3-VL-8B       &  8.6 & 16.1 & 19.2 & 25.7 & 32.6 & 39.6 \\
        BGE-M3       & + GPT-5.4           & 14.1 & 25.3 & 29.5 & 35.1 & 44.9 & 52.0 \\
        BGE-M3       & + Gemini-3.1-Pro    & 16.7 & 25.6 & 32.5 & 37.6 & 46.4 & 50.8 \\
        \midrule
        Qwen3-Emb-8B & raw question                & 20.4 & 40.0 & 48.5 & 57.0 & \textbf{69.4} & 75.6 \\
        Qwen3-Emb-8B & + Qwen3-VL-8B       & 17.4 & 30.1 & 37.5 & 46.5 & 57.5 & 63.6 \\
        Qwen3-Emb-8B & + GPT-5.4           & 26.7 & \textbf{43.3} & \textbf{52.4} & \textbf{60.2} & \textbf{69.4} & \textbf{76.3} \\
        Qwen3-Emb-8B & + Gemini-3.1-Pro    & \textbf{28.5} & 43.1 & 49.6 & 58.7 & 67.7 & 73.0 \\
        \bottomrule
    \end{tabular}
    
    \label{tab:retrieval_decomp}
    
\end{table}

We observe three main trends: 
(1) Directly using the raw user question as the search query performs poorly, especially with a lexical search backbone: BM25~\citep{10.1561/1500000019} achieves only $7.2\%$ R@10. This confirms a central mismatch in agentic video browsing: user questions often describe latent visual evidence needs, whereas video search operates over metadata-level surface forms such as titles, descriptions, and tags.
(2) Improving the search backbone substantially raises the search recall. Replacing BM25 with BGE-M3~\citep{chen2025m3embeddingmultilingualitymultifunctionalitymultigranularity} improves R@10 from $7.2\%$ to $23.2\%$, and Qwen3-Embedding-8B~\citep{zhang2025qwen3embeddingadvancingtext} further raises it to $48.5\%$. 
This suggests that when building a strong video search system, choosing a more powerful retriever is a first-order design choice.
(3) Translating user questions into searchable queries is still beneficial, but its effectiveness depends strongly on the quality of the planner. 
GPT-5.4 rewrites improve R@10 across all search backbone, with gains of $+16.7$, $+6.3$, and $+3.9$ points on BM25, BGE-M3, and Qwen3-Embedding-8B, respectively. 
By contrast, Qwen3-VL-8B rewrites hurt dense search, reducing Qwen3-Embedding-8B R@10 from $48.5\%$ to $37.5\%$.

\noindent\textbf{Reasoning axis.}\par\nopagebreak
\noindent To isolate the reasoning axis, we directly provide each model with the verified ground-truth video(s).

\begin{table}[H]
\centering
\begin{minipage}{.72\linewidth}

    \centering
    \caption{Reasoning accuracy when ground-truth videos are provided.}
    
    \footnotesize
    \setlength{\tabcolsep}{2pt}
    \begin{tabular*}{\linewidth}{@{\extracolsep{\fill}}lcccc@{}}
        \toprule
        \textbf{VLM} & \textbf{L1} & \textbf{L2} & \textbf{L3} & \textbf{All} \\
        \midrule
        Qwen3-VL-8B & 64.4 & 59.8 & 57.4 & 61.6 \\
        GPT-5.4 & 80.5 & 70.7 & \textbf{77.0} & 76.0 \\
        Gemini-3.1-Pro & \textbf{87.3} & \textbf{80.4} & 73.8 & \textbf{82.7} \\
        \bottomrule
    \end{tabular*}
    
    \label{tab:oracle_decomp}

\end{minipage}
\end{table}

In Table~\ref{tab:oracle_decomp}, video reasoning remains challenging even under oracle search: Qwen3-VL-8B achieves only $61.6\%$ overall accuracy, and all models show lower performance on implicit or multi-source cases than on explicit search cases. 
Stronger models raise the oracle ceiling, reaching $76.0\%$ for GPT-5.4 and $82.7\%$ for Gemini-3.1-Pro. 

\paragraph{Summary.}
The analysis exposes two coupled challenges: agents must first search the right video evidence, and then perform sufficiently strong fine-grained visual and cross-source reasoning over that evidence. 
Of the two, search is the more immediate bottleneck in open-web video setting: search is typically performed over metadata such as titles, descriptions, and tags, often with keyword-oriented matching. 
If an implicit visual evidence need is not translated into this searchable surface language, the answer-bearing video is never exposed to the agent.

\section{Method}
\label{sec:method}

\subsection{Overview}
\label{sec:method_overview}
\label{sec:iterative_browsing}
\begin{figure*}[htbp]
  \centering
   \includegraphics[width=1\linewidth]{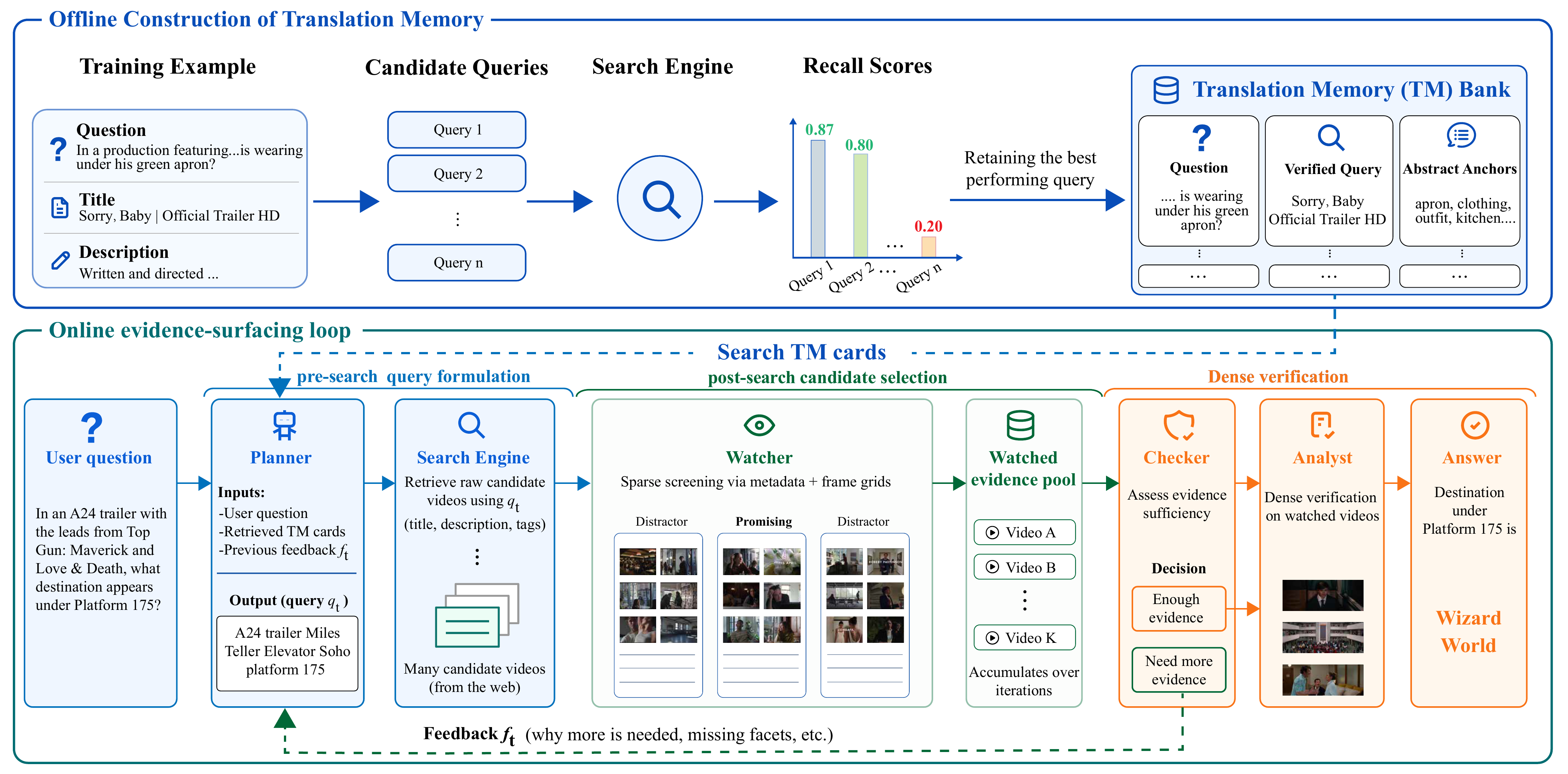}
   
   \caption{
        Overview of Video-Browser. The top panel constructs Translation Memory by selecting search-validated queries from training examples, while the bottom panel shows the online browsing loop, where the Planner generates searchable queries, the Watcher selects promising videos, and the Analyst performs dense visual verification.
    }
   
   \label{fig:pipeline}
\end{figure*}

Video-Browser is an agent system with four roles: a \emph{Planner} that formulates search queries, a \emph{Watcher} that screens returned candidates, a \emph{Checker} that decides when to stop browsing, and an \emph{Analyst} that produces the final answer.
As shown in Figure~\ref{fig:pipeline}, the central design principle is to identify promising videos before investing in costly, detailed video understanding:
the agent improves both \emph{pre-search query formulation} and \emph{post-search candidate selection} during browsing, and defers dense video inspection to a single final verification stage.

\paragraph{Iterative browsing.}
Given a user question $Q$, the Planner generates a search query $q_t$ at iteration $t$ using retrieved memory examples (Section~\ref{sec:memory}) and the browsing history $H_t$.
Here, $H_t$ contains the information available before iteration $t$,
including previous queries, candidate information, and Checker feedback.
The search engine returns a candidate set $V_t$, and the Watcher screens these videos using sparse visual previews, selecting a small subset $\hat{V}_t \subseteq V_t$.
Selected videos are accumulated across iterations into an evidence pool
$\hat{V}_{1:t} = \bigcup_{\tau=1}^{t} \hat{V}_\tau$.
The Checker then assesses whether this pool is sufficient to answer the question, based on the question, the browsing history, and the metadata and sparse previews of the accumulated videos:
\begin{equation}
    \mathrm{Checker}(Q, H_t, \hat{V}_{1:t})
    \rightarrow (d_t, f_t),
    \label{eq:checker}
\end{equation}
where $d_t \in \{\textsc{halt}, \textsc{continue}\}$ is the stopping decision and $f_t$ is textual feedback describing the missing evidence.
If $d_t = \textsc{continue}$, the feedback $f_t$ is appended to the history and guides the Planner's next query. Browsing ends when the Checker halts or the maximum number of iterations $T_{\max}$ is reached.

\paragraph{Final verification.}
After browsing ends, the Analyst examines denser frame samples and timestamped transcripts from the accumulated pool $\hat{V}_{1:T}$ to verify the evidence and generate the final answer.
This separates inexpensive candidate screening during browsing from detailed video inspection at the end.
The Planner, Checker, and Analyst are standard LLM roles instantiated with prompts (Appendix~\ref{agent_prompts}). The following subsections describe the two novel components that address the search bottleneck: Translation Memory for pre-search query formulation and the sparse Watcher for post-search candidate selection.

\subsection{Translation Memory}
\label{sec:memory}
User questions often refer to visual content in videos, whereas video search relies mainly on textual metadata.
Translation Memory provides past search examples for the Planner to draw on, helping it turn implicit user questions into search-engine-friendly queries.

\paragraph{Memory construction.}
For each training question, we provide a LLM with the question and the metadata of its paired target video.
The LLM generates several candidate search queries together with salient entities and keywords that may help identify the video.
We then execute each candidate query on the search engine and record the rank at which the target video appears in the returned results.
Among the candidates, we select the query that ranks the target video highest, and mark it as \emph{verified} if the target appears within the top-$K$ results ($K{=}3$ in our build).
If no candidate places the target within the top-$K$, we ask the LLM to generate a second set of queries with feedback.
Each successful question yields a verified memory card storing the training question, the verified search query, and several salient search terms.
If the retry also fails, we retain an \emph{unverified} card that stores only the question and its search-term hints.
All resulting cards form the Translation Memory $\mathcal{M}$.
Full construction details are provided in Appendix~\ref{app:memory_construction}.

\paragraph{Memory-guided query formulation.}
Given a new question $Q$, we retrieve the top-$k_m$ memory cards whose stored questions are most semantically similar to $Q$ under embedding-based cosine similarity.
We denote these retrieved examples by $\mathcal{M}_Q$ and provide them to the Planner together with the current browsing history:
\begin{equation}
    \mathrm{Planner}(Q, H_t, \mathcal{M}_Q)
    \rightarrow q_t .
    \label{eq:planner_query}
\end{equation}
The Planner draws on these examples to phrase new searches by analogy, using their query structures and search terms as references when formulating the next query $q_t$.

\subsection{Sparse Watcher}
\label{sec:watcher}

Even with improved search queries, the returned videos may have relevant-looking metadata but contain visual content unrelated to what the user is looking for.
Inspecting every returned video in detail is expensive.
We therefore introduce the Watcher, which uses sparse visual previews to quickly screen search results before further analysis.

At iteration $t$, the search engine returns a set of candidate videos $V_t$.
Each candidate is represented by its textual metadata and a small number of frames uniformly sampled across the video, arranged into timestamped grids (two $3{\times}3$ grids per video in our default setting).
This gives the Watcher a sparse visual preview of each candidate's actual content at a small, fixed cost per video.

The Watcher uses a multimodal LLM to jointly examine these visual previews and textual metadata in the context of the user question, current search query, and browsing history:
\begin{equation}
    \mathrm{Watcher}(Q, q_t, H_t, V_t)
    \rightarrow \hat{V}_t ,
    \label{eq:watcher}
\end{equation}
where $\hat{V}_t \subseteq V_t$ denotes the top-$k_w$ candidates selected by the Watcher ($k_w{=}1$ by default).

This allows the agent to use actual video content when distinguishing visually relevant candidates from metadata-similar distractors, without densely processing every returned video.
The selected videos join the accumulated evidence pool $\hat{V}_{1:t}$, which the Checker assesses at each iteration and the Analyst inspects in detail at the end.

\section{Experiments}
\label{sec:experiments}

In this section, we evaluate Video-Browser on the Video-BrowseComp benchmark.
Our experiments are organized around four research questions:
\textbf{RQ1 (Performance):} How does Video-Browser compare with ReAct~\citep{yao2023reactsynergizingreasoningacting} search agents and existing video-understanding agents?
\textbf{RQ2 (Translation Memory):} Does verified Translation Memory improve query formulation?
\textbf{RQ3 (Watcher):} Does sparse visual candidate selection improve post-search evidence selection?
\textbf{RQ4 (Ablation):} How much do Translation Memory, Watcher, and input modalities contribute to end-to-end browsing accuracy?
We use a fixed BM25 index for all offline experiments to approximate the text-based search interface available on open-web video platforms, where an agent faces a black-box search interface and cannot use dense retrievers to index an unseen corpus.
Full experimental settings are provided in Appendix~\ref{app:experimental_settings}.

\subsection{Metrics}

We report Overall Accuracy (OA) as the primary metric, following BrowseComp~\citep{wei2025browsecompsimplechallengingbenchmark}. 
An LLM judge~\citep{zheng2023judgingllmasajudgemtbenchchatbot} evaluates semantic equivalence between predictions and ground-truth answers. 
We also report Calibration Error (CE), where lower values indicate better alignment between self-reported confidence and empirical accuracy. 
Metric and prompts details are provided in Appendix~\ref{sec:metric_details} and Appendix~\ref{eval_prompts}.

\subsection{Baselines}
\label{sec:baselines}

We compare Video-Browser with two baseline families. 
First, ReAct search agents use Qwen3-VL-8B~\citep{yang2025qwen3technicalreport}, GPT-5.4\footnote{\url{https://developers.openai.com/api/docs/models/gpt-5.4}}, and Gemini-3.1-Pro\footnote{\url{https://deepmind.google/models/model-cards/gemini-3-1-pro}} with the same tools as Video-Browser. 
Second, video agents include VideoAgent~\citep{wang2024videoagentlongformvideounderstanding}, VideoTree~\citep{wang2025videotreeadaptivetreebasedvideo}, and DVD~\citep{zhang2025deepvideodiscoveryagentic}, adapted by feeding candidate videos from the original search stack with user question.

\subsection{Performance}
\label{sec:performance}

\begin{table*}[t]
    \centering
    \footnotesize
    
    \setlength{\tabcolsep}{2.6pt}
    \caption{Offline-track performance comparison across agent baselines. Recall denotes cumulative recall over candidates retrieved during the browsing loop. We report overall, difficulty-level, and genre-wise accuracy, with calibration error. \textbf{Bold} marks the best result.}
    
    \begin{tabular}{l c c ccc c c cccccccc}
        \toprule
        \multirow{2}{*}{\textbf{Model}} & \textbf{OA} & \textbf{Recall} & \multicolumn{3}{c}{\textbf{Difficulty}} & \textbf{CE} & & \multicolumn{8}{c}{\textbf{Genre Accuracy (\%)}} \\
        \cmidrule(lr){4-6} \cmidrule(lr){9-16}
        & \textbf{(\%)} & \textbf{(\%)} & \textbf{L1} & \textbf{L2} & \textbf{L3} & \textbf{(\%)} & & \textbf{Doc} & \textbf{Edu} & \textbf{Film} & \textbf{Game} & \textbf{Music} & \textbf{Sport} & \textbf{TV} & \textbf{Vid} \\
        \midrule
        \rowcolor[gray]{.95} \multicolumn{16}{l}{\textit{ReAct Search Agent}} \\
        Qwen3-VL-8B & 11.6 & 23.1 & 14.6 & 8.2 & 11.5 & 69.7 & & \heat{16.2} & \heat{7.8} & \heat{12.0} & \heat{1.8} & \heat{21.5} & \heat{9.0} & \heat{11.3} & \heat{13.1} \\
        GPT-5.4 & 27.8 & 41.8 & 34.6 & 20.7 & 26.2 & 21.7 & & \heat{35.1} & \heat{21.6} & \heat{30.0} & \heat{8.8} & \heat{43.1} & \heat{25.4} & \heat{32.3} & \heat{26.2} \\
        Gemini-3.1-Pro & 39.8 & 43.6 & 46.3 & \textbf{35.9} & 29.5 & 21.3 & & \textbf{\heat{48.6}} & \textbf{\heat{47.1}} & \heat{36.0} & \heat{26.3} & \textbf{\heat{61.5}} & \heat{25.4} & \heat{40.3} & \textbf{\heat{36.1}} \\
        \midrule
        \rowcolor[gray]{.95} \multicolumn{16}{l}{\textit{Video Agent}} \\
        VideoAgent & 5.3 & - & 3.9 & 7.6 & 3.3 & \textbf{14.3} & & \heat{18.9} & \heat{13.7} & \heat{2.0} & \heat{3.5} & \heat{3.1} & \heat{3.0} & \heat{1.6} & \heat{3.3} \\
        VideoTree & 8.0 & - & 5.9 & 9.2 & 11.5 & 16.5 & & \heat{27.0} & \heat{21.6} & \heat{6.0} & \heat{7.0} & \heat{3.1} & \heat{4.5} & \heat{1.6} & \heat{3.3} \\
        DVD & 17.8 & - & 19.5 & 17.4 & 13.1 & 39.9 & & \heat{21.6} & \heat{31.4} & \heat{12.0} & \heat{15.8} & \heat{16.9} & \heat{14.9} & \heat{14.5} & \heat{18.0} \\
        \midrule
        \rowcolor[gray]{.95} \multicolumn{16}{l}{\textit{Video-Browser (Ours)}} \\
        Qwen3-VL-8B & 27.3 & 46.2 & 34.6 & 20.1 & 24.6 & 56.8 & & \heat{16.2} & \heat{31.4} & \heat{26.0} & \heat{21.1} & \heat{38.5} & \heat{19.4} & \heat{35.5} & \heat{26.2} \\
        Gemini-3.1-Pro & \textbf{45.1} & \textbf{46.4} & \textbf{54.6} & 35.3 & \textbf{42.6} & 39.0 & & \heat{45.9} & \heat{45.1} & \textbf{\heat{40.0}} & \textbf{\heat{38.6}} & \heat{52.3} & \textbf{\heat{53.7}} & \textbf{\heat{46.8}} & \textbf{\heat{36.1}} \\
        \bottomrule
    \end{tabular}
    
    \label{tab:main_results_v1}
\end{table*}

Table~\ref{tab:main_results_v1} reports the main offline-track results.
Video-Browser achieves the best OA, reaching $27.3\%$ with Qwen3-VL-8B and $45.1\%$ with Gemini-3.1-Pro, improving over the corresponding ReAct search agents by $+15.7$ and $+5.3$ points.
Under the shared Qwen3-VL-8B backbone, the strongest adapted video-agent baseline, DVD, reaches $17.8\%$ OA, suggesting that long-form video reasoning alone does not solve search-heavy browsing.
Video-Browser also improves cumulative recall, reaching $46.2\%$ with Qwen3-VL-8B and $46.4\%$ with Gemini-3.1-Pro, with clear gains on both L1 and L3.

\subsection{Translation Memory}
\label{sec:query_phrasing}

We evaluate whether Translation Memory improves the Planner's query formulation.
For each question, we compare query generation without memory against generation with retrieved memory cards, and evaluate the resulting queries using one-shot Recall@10 across three search methods.

\begin{table}[H]
    \raggedright 
    
    \setlength{\tabcolsep}{4pt} 
    \scriptsize 

    \noindent
    \begin{minipage}[t]{0.54\linewidth}
        \vspace{0pt}
        \captionof{table}{One-shot search Recall@10 on Video-BrowseComp under different memory variants.}
        \label{tab:memory_variant_recall}
        
        \begin{tabular}{l l ccc}
            \toprule
            \textbf{Rewriter} & \textbf{Memory} & \textbf{BM25} & \textbf{BGE-M3} & \textbf{Qwen3-E} \\
            \midrule
            \multirow{2}{*}{Qwen3-VL-8B}
            & None
            & 15.5
            & 19.2
            & 37.5 \\

            & $\mathcal{M}$
            & \textbf{17.9}~\pos{2.4}
            & \textbf{21.7}~\pos{2.5}
            & \textbf{39.4}~\pos{1.9} \\

            \midrule
            \multirow{2}{*}{Gemini-3.1-Pro}
            & None
            & 29.9
            & 32.5
            & 49.6 \\

            & $\mathcal{M}$
            & \textbf{38.2}~\pos{8.3}
            & \textbf{40.2}~\pos{7.7}
            & \textbf{57.8}~\pos{8.2} \\

            \bottomrule
        \end{tabular}
    \end{minipage}\hfill
    \begin{minipage}[t]{0.40\linewidth}
        \vspace{0pt}
        \centering
        \includegraphics[width=\linewidth]{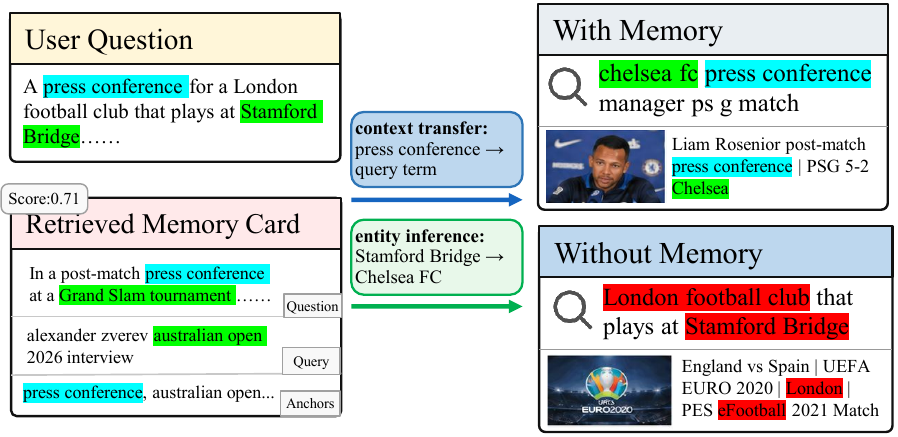}
        
        \captionof{figure}{Case figure for memory query.}
        \label{fig:memory_query}
    \end{minipage}

\end{table}

As shown in Table~\ref{tab:memory_variant_recall}, Translation Memory consistently improves Recall@10 across both planners and all three search backbones.
For Qwen3-VL-8B, the gains are modest but consistent, improving BM25, BGE-M3, and Qwen3-E by $+2.4$, $+2.5$, and $+1.9$ points, respectively.
For Gemini-3.1-Pro, the gains are substantially larger, reaching $+8.3$, $+7.7$, and $+8.2$ points.
These results show that past search experience provides a useful prior for formulating queries for new video-search questions.

Figure~\ref{fig:memory_query} further illustrates how this improvement arises.
Given a new question, the Planner retrieves a similar past question together with its search-validated query and uses this example to phrase the new query by analogy.
This can turn an implicit, visually grounded question into a more searchable query that better surfaces the target video.

\subsection{Watcher Analysis}
\label{sec:watcher_analysis}

We further evaluate whether the Watcher improves candidate selection after search.
We compare the Watcher against a metadata-only LLM Reranker.
Given the same BM25 Top-10 candidate pool, the LLM Reranker uses the same LLM to rank candidates based only on their textual metadata.

\begin{table}[H]
    \centering
    
    \setlength{\tabcolsep}{6pt}
    \footnotesize
    \caption{
        Post-search reranking on BM25 Top-10 results using Qwen3-VL-8B. 
        Top-1 Hit is measured over all questions; Cond. Top-1 and Cond. MRR@10 are computed only when the seed video is present in the BM25 Top-10 pool. 
        \textbf{Bold} marks the best result.
    }
    
    \begin{tabular}{lcccccc}
        \toprule
        \textbf{Method} 
        & \textbf{Top-1 Hit} 
        & \textbf{Cond. Top-1}
        & \textbf{Cond. MRR@10}
        & \textbf{L1 Top-1}
        & \textbf{L2 Top-1}
        & \textbf{L3 Top-1} \\
        \midrule
        Search Engine
        & 6.0
        & -- 
        & -- 
        & 7.3
        & 2.2
        & 13.1 \\
        
        LLM Reranker
        & 12.0
        & 57.4
        & 73.5
        & 11.2
        & 7.1
        & 29.5 \\
        
        Watcher
        & \textbf{15.6}~\pos{3.6}
        & \textbf{75.3}~\pos{17.9}
        & \textbf{83.6}~\pos{10.1}
        & \textbf{15.6}~\pos{4.4}
        & \textbf{9.8}~\pos{2.7}
        & \textbf{32.8}~\pos{3.3} \\
        
        \bottomrule
    \end{tabular}
    
    \label{tab:stage1_selector_ablation}
\end{table}

As shown in Table~\ref{tab:stage1_selector_ablation}, directly taking the search engine's top-ranked result is insufficient, achieving only 6.0\% Top-1 hit rate.
The LLM Reranker improves the Top-1 hit rate to 12.0\%, indicating that language models can use metadata to filter some irrelevant candidates.
However, the Watcher further improves Top-1 hit rate to 15.6\% and raises conditional accuracy from 57.4\% to 75.3\%.
This shows that sparse visual previews provide useful evidence for distinguishing the seed video from metadata-similar hard negatives.

The improvement is consistent across all three difficulty levels.
Compared with the metadata-only LLM Reranker, the Watcher improves Top-1 hit rate by 4.4 points on Level 1, 2.7 points on Level 2, and 3.3 points on Level 3.
These results suggest that post-search visual screening is complementary to query formulation: Translation Memory helps search better candidate pools, while the Watcher prevents visually irrelevant candidates from being passed to the final Analyst.

\subsection{Ablation Study}
\label{sec:ablation}

Table~\ref{tab:ablation} analyzes the contribution of the two evidence-surfacing mechanisms and the input modalities.
The component rows are cumulative: the baseline uses the same browsing pipeline without Translation Memory and Watcher, then we add Translation Memory and Watcher sequentially.
The modality rows remove either transcript or visual input from the full system.

\begin{table}[H]
\centering
\begin{minipage}{.72\linewidth}

    \centering
    \footnotesize
    \caption{Component and modality ablations.}
    
    \setlength{\tabcolsep}{2pt}
    \begin{tabular*}{\linewidth}{@{\extracolsep{\fill}}lcccc@{}}
        \toprule
        \textbf{Setting} & \textbf{OA} & \textbf{CE} & \textbf{Recall} & \textbf{Watched} \\
        \midrule
        Baseline & 22.2 & 66.9 & 42.9 & 34.2 \\
        \midrule
        $+$ Memory  & 25.1 & 64.3 & 46.2 & 34.4 \\
        $+$ Watcher & 27.3 & 56.8 & 46.2 & 36.2 \\
        \midrule
        w/o Transcript & 19.8 & 65.1 & 45.3 & 34.9 \\
        w/o Visual     & 9.8  & 71.4 & 46.0 & 34.9 \\
        \bottomrule
    \end{tabular*}
    \label{tab:ablation}

\end{minipage}
\end{table}

\noindent\textbf{Components.}
The component rows show that both evidence-surfacing modules contribute to end-to-end performance.
Adding Translation Memory improves OA from $22.2\%$ to $25.1\%$ and Recall from $42.9\%$ to $46.2\%$, supporting its role in pre-search query formulation.
Adding the Watcher further improves OA to $27.3\%$ and watched from $34.4\%$ to $36.2\%$, suggesting that the gain is driven primarily by better visual candidate selection rather than by a large increase in the number of inspected videos.

\noindent\textbf{Modality.}
Both transcript and visual input contribute to final performance, but visual evidence is more critical.
Removing transcripts reduces OA from $27.3\%$ to $19.8\%$, showing that transcripts provide useful semantic context.
Removing visual input causes a much larger drop to $9.8\%$ and worsens CE to $71.4\%$, confirming that textual evidence alone cannot replace fine-grained visual verification in Video-BrowseComp.

\section{Conclusion}

We formalized \textit{Agentic Video Browsing}, where an agent must search, inspect, and verify evidence from open-web videos, and introduced \textbf{Video-BrowseComp} as a reproducible benchmark infrastructure for this setting.
Our analysis shows that the key challenge is evidence surfacing: agents must search and retain answer-bearing videos before final visual reasoning can succeed.
We proposed \textbf{Video-Browser}, which addresses this challenge with \emph{Translation Memory} for pre-search query formulation and a \emph{Watcher} for post-search visual candidate selection.
Experiments show that Video-Browser improves over corresponding search-agent and video-agent baselines, demonstrating that search, candidate selection, and visual verification should be treated as coupled components in agentic video browsing.
We hope Video-BrowseComp and Video-Browser provide a foundation for future research on reliable and efficient video browsing agents.

\section*{AI Use Statement}
Generative AI tools were used in three parts of the research workflow.
First, they were used to assist with language editing and to improve the clarity and readability of the manuscript.
Second, they were used to support literature retrieval and discovery, including identifying potentially relevant prior work; all cited references and their descriptions were subsequently checked by the authors against the original sources.
Third, generative AI was used in the automatic construction of the 1,000-question training split used for Translation Memory. The generated samples were retained only after the consistency and retrieval-based filtering procedures described in Appendix~\ref{data_pipeline}. The 660-question evaluation set was human-annotated and independently cross-validated, rather than generated by AI. All AI-assisted outputs were reviewed and verified by the authors, who take full responsibility for the final content of the paper.

\section*{Ethics Statement}
This work involves annotation by consenting participants and research over publicly accessible web videos. We follow the Code of Ethics and take measures to address privacy, copyright, data provenance, and responsible release concerns. Detailed discussion of potential risks, safeguards, and asset documentation is provided in Appendix \ref{appendix:limitandsocietal}.

\section*{Reproducibility Statement}
We provide detailed dataset construction procedures, evaluation metrics, agent configurations, prompts, and implementation settings in the Appendix \ref{appendix:benchmark} and Appendix \ref{appendix:agent}. The offline evaluation uses a fixed video corpus and search stack to ensure reproducible candidate pools and search results. Anonymous code and benchmark resources are provided with the submission.

\newpage
\bibliography{references}
\bibliographystyle{plainnat}

\newpage
\appendix
\section*{Appendix}
\label{sec:appendix}

\noindent\framebox[\linewidth]{
    \begin{minipage}{0.95\linewidth}
        \vspace{0.2cm}
        \noindent\textbf{Contents}
        \begin{itemize}[leftmargin=1.5em, itemsep=1pt]
            \item \textbf{\S\ref{appendix:benchmark} Video-BrowseComp (Benchmark Details)}
            \begin{itemize}[label=$\circ$]
                \item \S\ref{sec:bench_samples} Benchmark Samples
                \item \S\ref{annotators} Annotators
                \item \S\ref{data_pipeline} Dataset Construction
                \item \S\ref{detailed_analysis} Detailed Dataset Statistics
                \item \S\ref{app:hard_negatives} Hard-Negative Pool Construction
                \item \S\ref{sec:metric_details} Evaluation Metrics Details
                \item \S\ref{eval_prompts} Evaluation Prompts
                \item \S\ref{online_videosearch} Online Video Search Performance
            \end{itemize}

            \item \textbf{\S\ref{appendix:agent} Video-Browser (Agent Details)}
            \begin{itemize}[label=$\circ$]
                \item \S\ref{app:experimental_settings} Experiment Settings
                \item \S\ref{app:memory_construction} Translation Memory Construction
                \item \S\ref{agent_prompts} Prompts
                \item \S\ref{failure_case} Failure Case Analysis
            \end{itemize}
            \item \textbf{\S\ref{appendix:limitandsocietal} Limitations and Societal Considerations}
            \begin{itemize}[label=$\circ$]
                \item \S\ref{sec:limitations} Limitations
                \item \S\ref{sec:broader_impacts} Broader Impacts
                \item \S\ref{sec:responsible_release} Responsible Release, Safeguards, and Asset Documentation
            \end{itemize}
        \end{itemize}
        \vspace{0.1cm}
    \end{minipage}
}
\vspace{0.5cm}

\section{Video-BrowseComp}
\label{appendix:benchmark}

\subsection{Benchmark Samples}
\label{sec:bench_samples}

\begin{figure*}[htbp]
  \centering
   \includegraphics[width=1\linewidth]{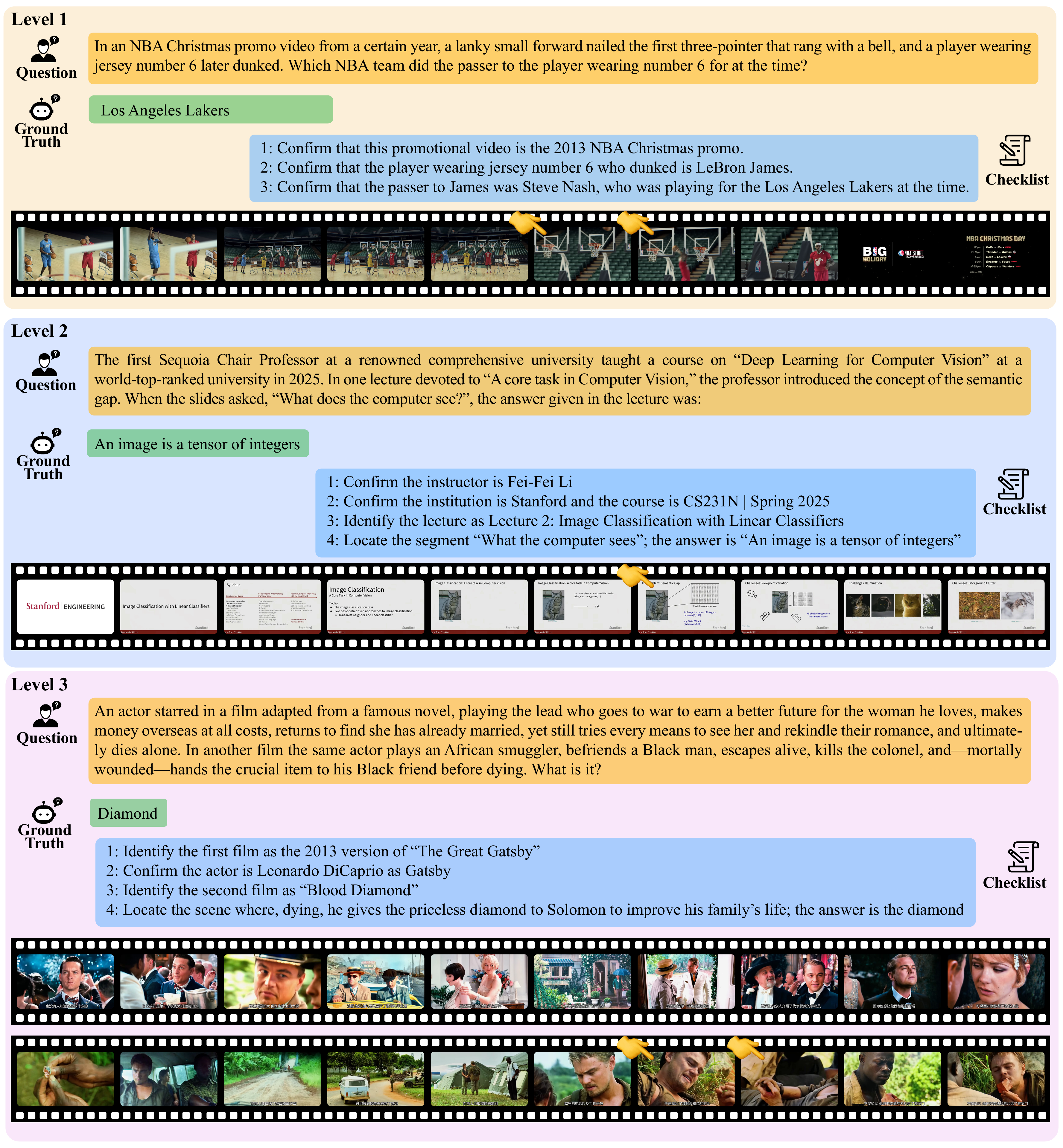}
   
   \caption{
   Benchmark samples.
   }
   
   \label{fig:data_samples}
\end{figure*}

We provide some benchmark examples in Figure \ref{fig:data_samples} for reference.

\subsection{Annotators}
\label{annotators}

To curate Video-BrowseComp, we recruited 8 student volunteers (holding Master's or PhD degrees) with extensive experience in LLM usage, web search, and video understanding. The construction process follows a rigorous annotation pipeline with two-stage quality control in Appendix \ref{data_pipeline}.

Informed consent was obtained from all participants prior to the study. They were informed that the collected data would be used for research purposes and released publicly.

\subsection{Dataset Construction}
\label{data_pipeline}

To make Video-BrowseComp evaluate agentic video browsing rather than general web search, 
we impose three construction criteria. First, each question must require visual evidence 
from videos. Second, the target evidence 
should be hard to find but easy to verify: questions may require indirect retrieval, 
cross-referencing, or temporal grounding, but answers are constrained to short objective 
strings such as entity names, colors, or counts. Third, each question must admit a unique 
answer grounded in observable spatio-temporal evidence rather than subjective interpretation.

\textbf{Annotation Pipeline.} We adopt a reverse construction strategy that begins with video exploration and culminates in complex query formulation. First, each annotator is assigned 2 video categories and encouraged to explore relevant topics through web browsing. After identifying candidate videos, annotators watch the complete video and design an initial question-answer pair focused on a specific clip. These simple questions are then expanded into complex queries following two principles: \textit{Indirect \& Compositional Queries:} Questions are augmented with compositional conditions to ensure the target video cannot be easily retrieved through a simple search. Annotators use indirect descriptions to reference video content rather than explicit keywords. \textit{Multi-Source Reasoning:} Level 1 and Level 2 samples can be derived from the above steps with single-video evidence. For Level 3 samples, annotators first identify a set of related videos covering the same topic or event, then design questions that require aggregating complementary information across these videos, such as comparing statistics from different matches, verifying conflicting claims across multiple sources, or tracing an event's progression through several recordings. Finally, annotators document the complete evidence chain and seed video URLs.

\textbf{Quality Control.} To ensure that each question has only one correct and unambiguous answer. We employ a cross-validation strategy among the annotators: each annotator answers questions annotated by others following the evidence chain and seed video URLs, and we compare their responses against the original ground-truth answers. If an alternative answer satisfies all task constraints but differs from the original, the sample is considered ambiguous and discarded.

\subsection{Detailed Dataset Statistics}
\label{detailed_analysis}

This section reports the per-genre and per-difficulty composition of Video-BrowseComp, the question / answer / evidence-chain length distributions, and the offline video pool that grounds all retrieval experiment.  

\paragraph{Evaluation Set Composition (660 questions).}
The 660-question evaluation set is partitioned into an \textit{online} track (210 questions, evaluated against the live web) and an \textit{offline} track (450 questions, evaluated against the offline pool). At the video-ID level, the training split is disjoint from the offline evaluation set and overlaps with only two videos in the online evaluation set. Table~\ref{tab:eval_split} reports the per-genre $\times$ per-difficulty cell counts for each track.

\paragraph{Training Set Composition (1{,}000 questions).}
The training split contains 1{,}000 question--answer pairs produced by an autonomous reverse-construction pipeline that is independent of the human-annotated eval set. The pipeline draws a seed video from the offline pool, prompts an LLM to summarize visual highlights and on-screen entities, and then generates a candidate question that the same LLM must be able to answer from the seed video alone. 
A generated sample is kept only if (i) the auto-judged answer matches a regenerated answer under a paraphrase of the same question, and (ii) the seed video can be re-retrieved via at least one BM25 query proposed by the LLM. 
The retained 1{,}000 rows form the source training split used for Translation Memory construction (Appendix~\ref{app:memory_construction}) and are balanced 500/500 across L1/L2 difficulties. Per-genre counts appear in the right block of Table~\ref{tab:eval_split}.

\paragraph{Question, Answer, and Evidence-Chain Statistics.}
Across the 660 eval questions, the question text averages 54.9 words (median 53; range 18--139). Answers are short and verifiable: 1.5 words on average (median 1, max 10) and 8.0 characters on average (median 6), consistent with the ``objective short string'' answer constraint of Section~\ref{data_pipeline}. The annotated evidence chain (the ordered list of retrieval-and-verification steps an annotator followed to confirm the answer) has 3.1 steps on average and is concentrated at length 3 (69\% of questions), with 21\% at length 4 and a 1\% tail at length 5 reserved for the hardest L3 multi-source cases. Mean chain length is essentially constant across L1, L2, and L3 (3.12, 3.12, 3.10 respectively).

\begin{table}[h]
    \centering
    \footnotesize
    \setlength{\tabcolsep}{4pt}
    \caption{Per-genre and per-difficulty composition of Video-BrowseComp. The eval set is split into offline (450) and online (210) tracks, the training set (1{,}000) is auto-generated.}
    \begin{tabular}{l ccc c ccc c cc c c c}
        \toprule
        \multirow{2}{*}{\textbf{Genre}}
        & \multicolumn{3}{c}{\textbf{Eval Offline (450)}}
        & & \multicolumn{3}{c}{\textbf{Eval Online (210)}}
        & & \multicolumn{2}{c}{\textbf{Train (1000)}}
        & & \multirow{2}{*}{\textbf{Eval All}}
        & \multirow{2}{*}{\textbf{Train All}} \\
        \cmidrule(lr){2-4} \cmidrule(lr){6-8} \cmidrule(lr){10-11}
        & L1 & L2 & L3 & & L1 & L2 & L3 & & L1 & L2 & & & \\
        \midrule
        Documentary           & 18 &  9 & 10 & &  2 &  8 &  0 & & 53 & 45 & &  47 &  98 \\
        Education \& Knowledge & 19 & 26 &  6 & & 23 &  2 &  0 & & 68 & 74 & &  76 & 142 \\
        Film                  & 21 & 17 & 12 & & 46 &  0 & 15 & & 67 & 56 & & 111 & 123 \\
        Games \& Esports      & 21 & 22 & 14 & & 13 &  9 &  0 & & 60 & 46 & &  79 & 106 \\
        Music \& Variety      & 32 & 28 &  5 & &  5 &  5 &  0 & & 35 & 53 & &  75 &  88 \\
        Sports                & 30 & 27 & 10 & & 10 & 31 &  4 & & 91 & 96 & & 112 & 187 \\
        TV Series             & 31 & 27 &  4 & & 16 &  2 &  1 & & 52 & 36 & &  81 &  88 \\
        Video \& Shorts       & 33 & 28 &  0 & & 10 &  5 &  3 & & 74 & 94 & &  79 & 168 \\
        \midrule
        \textbf{Total}        & \textbf{205} & \textbf{184} & \textbf{61}
                              & & \textbf{125} & \textbf{62} & \textbf{23}
                              & & \textbf{500} & \textbf{500}
                              & & \textbf{660} & \textbf{1000} \\
        \bottomrule
    \end{tabular}
    \vspace{2mm}
    \label{tab:eval_split}
\end{table}

\paragraph{Offline Video Pool Composition.}
The final hard-negative-augmented offline pool contains 12{,}455 unique videos, with metadata (title, description, channel, tags) frozen at cache time so that search results are reproducible. The pool totals \textbf{2{,}178\,hours} of footage, with a median length of 1.6\,minutes (mean 10.5; p25 = 32\,seconds, p75 = 8.2\,minutes, p95 = 55.6\,minutes; range from 2\,seconds to 34.8\,hours for a small tail of livestream archives). The distribution is heavy-tailed and YouTube-typical, dominated by short clips: 43.5\% of videos are under 1\,minute and 69.1\% are under 5 minutes, while only 4.3\% exceed one hour (Table~\ref{tab:duration_distribution}).

\begin{table}[h]
    \centering
    \footnotesize
    \setlength{\tabcolsep}{8pt}
    \caption{Video duration distribution over the 12{,}455 videos in the offline pool.}
    \begin{tabular}{l c c}
        \toprule
        \textbf{Length bucket} & \textbf{\# videos} & \textbf{\% of pool} \\
        \midrule
        $<$ 1\,minute    & 5{,}413 & 43.5\% \\
        1--5\,minutes    & 3{,}195 & 25.7\% \\
        5--10\,minutes   & 1{,}138 &  9.1\% \\
        10--30\,minutes  & 1{,}486 & 11.9\% \\
        30--60\,minutes  &    682 &  5.5\% \\
        $>$ 1\,hour      &    541 &  4.3\% \\
        \midrule
        \textbf{Total}   & \textbf{12{,}455} & \textbf{100.0\%} \\
        \bottomrule
    \end{tabular}
    \vspace{2mm}
    \label{tab:duration_distribution}
\end{table}

\subsection{Hard-Negative Pool Construction}
\label{app:hard_negatives}

The hard negatives are added to the base pool to deliberately stress post-search candidate selection: each one is a video that the BM25 index ranks within the top-10 for at least one evaluation question but that subsequent text and visual verification confirms does not contain the answer. Without such decoys, the BM25 top-10 is too lenient: the seed video is often the only topically relevant candidate, allowing a metadata-only LLM reranker to identify it with little need for visual evidence. We therefore construct hard negatives through a five-stage offline pipeline (Table~\ref{tab:hardneg_pipeline}) that combines broad topical search with a two-layer text-then-visual verification cascade.

\paragraph{Stage 1: Decoy query generation.}
For each of the 450 offline-track questions, we issue four planner-rewriter caches against the YouTube Data API. Three of the four caches are the same query rewrites used at inference time by Qwen3-VL-8B-Instruct, GPT-5.4, and Gemini-3.1-Pro (i.e., rewrites optimized to \textit{retrieve the answer}). The fourth is a dedicated \textit{decoy} cache produced by a separate Qwen3-VL-8B prompt that asks for three broader topical queries which (i) describe the question's domain rather than its specific entities, (ii) avoid quoted exact phrases or rare proper nouns, and (iii) avoid any words that appear in the gold answer. The decoy cache contributes 13{,}470 of the 19{,}895 (question,\,video,\,rewriter,\,rank) tuples returned by Stage 1 and is the dominant source of plausible-but-wrong candidates.

\paragraph{Stage 2: Prefilter.}
Stage 2 deduplicates the search results into 17{,}168 unique (question,\,video) candidate pairs and drops two contaminating classes: (i) videos already present in the base offline pool, and (ii) videos that are an annotated seed/source/positive for \emph{any} question in the full benchmark, so that a hard negative for question $i$ cannot be a ground-truth video for question $j$. For each surviving pair, we also compute a \texttt{rewriter\_overlap\_score}, defined as the number of the four rewriter caches that surfaced this video for this question, which serves as a multi-rewriter convergence signal in Stage 5.

\paragraph{Stage 3: Layer-1 text verification.}
A text-only Qwen3-VL-8B-Instruct reads the question, the candidate's title / description / channel / tags, and, when available, the YouTube auto-subtitle transcript, and emits a structured verdict $\in \{\textsc{Yes}, \textsc{Uncertain}, \textsc{No}\}$ on whether the video contains the answer, together with a topical-relevance label $\in \{\textsc{High}, \textsc{Mid}, \textsc{Low}\}$. Across the 17{,}168 candidates, the verdict distribution is \textsc{No}~$=15{,}900$ ($92.6\%$), \textsc{Uncertain}~$=1{,}032$ ($6.0\%$), and \textsc{Yes}~$=236$ ($1.4\%$); the relevance distribution is \textsc{High}~$=3{,}607$, \textsc{Mid}~$=6{,}544$, and \textsc{Low}~$=7{,}017$.

We retain only candidates assigned a \textsc{No} verdict with \textsc{High} or \textsc{Mid} topical relevance. \textsc{Yes} candidates are excluded as potential covert positives, \textsc{Uncertain} candidates are excluded because they cannot be confidently treated as negatives, and \textsc{Low}-relevance candidates are removed as off-topic. Applying this criterion leaves 8{,}883 candidates. We sort them by $(\texttt{rewriter\_overlap\_score}, -\texttt{yt\_rank\_min})$ and shortlist up to three candidates per question for the more expensive Layer-2 verification, yielding 707 shortlisted candidates.

\paragraph{Stage 4: Layer-2 visual verification.}
Each of the 707 shortlisted candidates is downloaded, and 16 frames are uniformly sampled at temporal midpoints. The same Qwen3-VL-8B-Instruct re-judges the question against the candidate's title, description, and 16 sparse frames under the same verdict / relevance schema as Layer 1. Across these 707 candidates, the verdict distribution is \textsc{No}~$=666$ ($94.2\%$), \textsc{Yes}~$=40$ ($5.7\%$), and \textsc{Uncertain}~$=1$; the relevance distribution is \textsc{High}~$=429$, \textsc{Mid}~$=161$, and \textsc{Low}~$=117$.

The visual verification stage catches a non-trivial number of metadata-only false negatives: 40 candidates that passed Layer 1 as \textsc{No} are reassigned \textsc{Yes} once frames are inspected, while 117 candidates are assigned \textsc{Low} visual relevance despite having passed the text-based relevance filter. Applying the same eligibility criterion as above---a \textsc{No} verdict with \textsc{High} or \textsc{Mid} relevance---leaves 549 candidates for final selection.

\paragraph{Stage 5: Selection and integration.}
For each question, we rank the surviving candidates by 
\[
\begin{aligned}
(&\texttt{rewriter\_overlap\_score},\ \texttt{best\_relevance\_across\_layers},\\
 &\texttt{layer2\_verdict},\ \texttt{yt\_rank\_min})
\end{aligned}
\]
 in decreasing priority and keep at most the top three. Only candidates that satisfy the negative-verification criterion---a \textsc{No} verdict with \textsc{High}/\textsc{Mid} relevance---are eligible for final selection. This produces 522 (question,\,hard-negative) pairs, which are merged into the base offline pool. Their metadata is frozen alongside the existing cache so that subsequent search results remain reproducible. The resulting hard-negative-augmented pool is the 12{,}455-video offline pool reported in Appendix~A.4. Across the 450 offline questions, the 522 retained pairs correspond to an average of 1.16 hard-negative assignments per question.

\begin{table}[h]
    \centering
    \footnotesize
    \setlength{\tabcolsep}{4pt}
    \caption{Hard-negative pipeline funnel. Candidate counts denote the number surviving each filtering stage. Layer-2 evaluates 707 candidates shortlisted from the 8{,}883 Layer-1 survivors, of which 549 pass visual verification.}
    \begin{tabular*}{\linewidth}{@{\extracolsep{\fill}}p{0.41\linewidth} c p{0.41\linewidth}@{}}
        \toprule
        \textbf{Stage} & \textbf{Candidates} & \textbf{Filter applied} \\
        \midrule
        1. Decoy search (4 rewriters $\times$ 450 q.) 
        & 19{,}895 
        & --- \\

        2. Prefilter (dedup, drop pool, drop GT) 
        & 17{,}168 
        & remove existing pool / any-question GT \\

        3. Layer-1 text verify (Qwen3-VL-8B) 
        & 8{,}883 
        & keep \textsc{No} with \textsc{High}/\textsc{Mid} relevance \\

        4. Layer-2 visual verify (707 shortlisted) 
        & 549 
        & keep \textsc{No} with \textsc{High}/\textsc{Mid} relevance \\

        5. Final selection (top-3 merged into pool) 
        & \textbf{522} 
        & per-question cap and selection ranking \\
        \bottomrule
    \end{tabular*}
    \vspace{2mm}
    \label{tab:hardneg_pipeline}
\end{table}

\subsection{Evaluation Metrics Details}
\label{sec:metric_details}

\subsubsection{LLM-based Accuracy Judge}
Since the reference answers in Video-BrowseComp are designed to be short and verifiable (e.g., specific entities, colors, or counts), rigid string matching often fails due to minor formatting differences. We utilize gpt-5-mini-2025-08-07\footnote{\url{https://openai.com/index/introducing-gpt-5/}} as the automated judge. The judge is prompted to verify if the prediction conveys the same factual information as the ground truth. The specific prompt used for evaluation is shown in Figure \ref{fig:eval_prompts}.

\subsubsection{Calibration Error (CE)}
To calculate CE, we first prompt the model to provide a confidence score $p \in [0, 1]$ alongside its final answer. We partition the predictions into $B=5$ equally spaced bins (i.e., $[0, 0.2), \dots, [0.8, 1.0]$). The CE is calculated as the weighted average of the absolute difference between the accuracy and the average confidence within each bin:

\begin{equation}
    \text{CE} = \sum_{i=1}^{B} \frac{n_i}{N} | \text{acc}(i) - \text{conf}(i) |
\end{equation}

Where $N$ is the total number of samples, $n_i$ is the number of samples in the $i$-th bin, $\text{acc}(i)$ is the empirical accuracy of the samples in bin $i$, and $\text{conf}(i)$ is the average predicted confidence of the samples in that bin.

\subsubsection{Cumulative Recall}
\label{sec:cumulative_recall}
Cumulative Recall measures whether an agent's browsing loop ever surfaces an answer-bearing video into a candidate pool that the downstream Watcher / Analyst can inspect; it is the second column of Table~\ref{tab:main_results_v1}. Let $T_i$ denote the number of search loops executed for question $Q_i$, with $T_i \leq T_{\max}$, and let $V_t^{(i)}$ denote the BM25 candidate set returned at loop $t \in \{1, \ldots, T_i\}$. Let $\mathcal{G}_i$ denote the set of annotated ground-truth videos for $Q_i$; $\lvert\mathcal{G}_i\rvert=1$ for L1/L2 questions, while L3 questions may contain multiple ground-truth videos.
For L3 questions, this metric uses an any-hit criterion: the question is counted as retrieved if at least one video in $\mathcal{G}_i$ is surfaced, rather than requiring recovery of the complete multi-source evidence chain.
We pool all candidates surfaced across loops,
$V_{1:T_i}^{(i)} = \bigcup_{t=1}^{T_i} V_t^{(i)}$,
and report the fraction of questions for which at least one annotated ground-truth video appears anywhere in this union:
\begin{equation}
    \mathrm{Recall}_{\mathrm{cum}}
    =
    \frac{1}{N}
    \sum_{i=1}^{N}
    \mathbf{1}
    \left[
    \mathcal{G}_i
    \cap
    \left(
    \bigcup_{t=1}^{T_i} V_t^{(i)}
    \right)
    \neq \emptyset
    \right].
\end{equation}
This metric differs from conventional R@$k$ in two ways. First, R@$k$ measures single-shot retrieval quality for a fixed query, whereas Cumulative Recall measures the agent's multi-step ability to formulate queries that collectively surface at least one annotated ground-truth evidence video. Second, R@$k$ truncates results at a fixed rank $k$ for one query, whereas Cumulative Recall takes the union over all queries issued during the browsing loop. The number of loops is determined by the agent's stopping behavior and capped at $T_{\max}=3$ in our default configuration. Thus, Cumulative Recall measures whether the browsing trajectory exposes at least one verified evidence source that can be inspected by the downstream verifier.

\subsection{Evaluation Prompts}
\label{eval_prompts}

We provide the prompts used in evaluation and judge process at Figure \ref{fig:eval_prompts}.

\par\begingroup
\captionsetup{hypcap=false}

    \begin{tcolorbox}[breakable,colback=gray!10, colframe=gray!50, title={Evaluation Prompt}]
        Your response should be in the following JSON format:
\begin{verbatim}
{
    "Explanation": "your explanation for your final answer",
    "Answer": "your succinct, final answer",
    "Confidence": "your confidence score between 0% and 100% 
    for your answer"
}
\end{verbatim}
    \end{tcolorbox}
    \vspace{0.2cm} 
    \begin{tcolorbox}[breakable,colback=gray!10, colframe=gray!50, title={Judge Prompt}]
        \begin{verbatim}
Question: {question}

Ground Truth: {ground_truth}

Model Prediction: {prediction}

Evaluate if the Prediction matches the Ground Truth.
Return JSON: {{"is_correct": true}} or {{"is_correct": false}}
\end{verbatim}

    \end{tcolorbox}
    \captionof{figure}{Full prompts used for evaluation and judging.}
    \label{fig:eval_prompts}
\par\endgroup

\subsection{Online Video Search Performance}
\label{online_videosearch}

The offline-track results in Table~\ref{tab:main_results_v1} fix the candidate pool to ensure reproducibility, but they cannot speak to how each agent behaves when allowed to query the live web. We therefore include a complementary online-track evaluation on the 210 online questions of Video-BrowseComp (Appendix~\ref{detailed_analysis}), where every agent issues real-time YouTube searches without any pre-cached candidate pool. We compare two baseline families with web search enabled --- closed-source agents (Gemini-2.5-Flash, Gemini-2.5-Pro~\citep{comanici2025gemini25pushingfrontier}, GPT-5.1\footnote{\url{https://openai.com/index/introducing-gpt-5/}}) and the deep-research agent (o4-mini-deep-research\footnote{\url{https://openai.com/index/introducing-deep-research/}}) --- against two configurations of Video-Browser using Qwen3-VL-8B-Instruct and GPT-5.1. As shown in Table~\ref{tab:main_results}, Video-Browser with GPT-5.1 achieves the highest overall accuracy ($26.19\%$), providing complementary evidence that the gains we observe on the offline track can extend to live-web browsing; in contrast the deep-research agent leads on the harder L2 / L3 sub-splits but at the cost of substantially worse calibration. These numbers were collected during an earlier benchmarking pass against the prior generation of frontier APIs (Gemini-2.5, GPT-5.1) and are reported here as preliminary online evidence; a refresh against the GPT-5.4 / Gemini-3.1-Pro endpoints used in the offline-track main results is left for future work.

\begin{table*}[t]
    \centering
    \footnotesize
    
    \setlength{\tabcolsep}{2.6pt}
    \caption{Comprehensive results with a heatmap on Genre Accuracy. \textbf{Bold} indicates best performance. * means model has web search ability.}
    \resizebox{\textwidth}{!}{%
    \begin{tabular}{l c ccc c cc cccccccc}
        \toprule
        \multirow{2}{*}{\textbf{Model}} & \textbf{OA} & \multicolumn{3}{c}{\textbf{Difficulty}} & \textbf{CE} & & \multicolumn{8}{c}{\textbf{Genre Accuracy (\%)}} \\
        \cmidrule(lr){3-5} \cmidrule(lr){8-15}
        & \textbf{(\%)} & \textbf{L1} & \textbf{L2} & \textbf{L3} & \textbf{(\%)} & & \textbf{Doc} & \textbf{Edu} & \textbf{Film} & \textbf{Game} & \textbf{Music} & \textbf{Sport} & \textbf{TV} & \textbf{Vid} \\
        \midrule
        \rowcolor[gray]{.95} \multicolumn{15}{l}{\textit{Close Source Search Agent}} \\
        Gemini-2.5-Flash* & 20.95 & 32.80 & 4.84 & 0.00 & 35.98 & & \heat{0.00} & \heat{36.00} & \heat{29.51} & \heat{4.55} & \heat{10.00} & \heat{11.11} & \heat{36.84} & \heat{16.67} \\
        Gemini-2.5-Pro* & 23.81 & \textbf{37.60} & 4.84 & 0.00 & 31.45 & & \textbf{\heat{20.00}} & \heat{32.00} & \textbf{\heat{31.15}} & \heat{9.09} & \heat{10.00} & \heat{8.89} & \textbf{\heat{57.89}} & \heat{16.67} \\
        GPT-5.1* & 15.24 & 21.60 & 6.45 & 4.35 & \textbf{30.20} & & \heat{0.00} & \heat{8.00} & \heat{21.31} & \heat{18.18} & \heat{10.00} & \heat{8.89} & \heat{21.05} & \heat{22.22} \\
        o4-mini-deep-research & 22.86 & 30.40 & \textbf{12.90} & \textbf{8.70} & 42.55 & & \heat{10.00} & \heat{28.00} & \heat{29.51} & \textbf{\heat{27.27}} & \heat{10.00} & \heat{17.78} & \heat{21.05} & \heat{16.67} \\
        \midrule
        \rowcolor[gray]{.95} \multicolumn{15}{l}{\textit{Video-Browser}} \\
        Video-Browser (Qwen3) & 19.05 & 25.60 & 9.68 & \textbf{8.70} & 69.81 & & 0.00 & 36.00 & 22.95 & 4.55 & 10.00 & 11.11 & 15.79 & 38.89 \\
        Video-Browser (GPT-5.1) & \textbf{26.19} & \textbf{37.60} & 11.29 & 4.35 & 44.60 & & \heat{10.00} & \textbf{\heat{40.00}} & \textbf{\heat{31.15}} & \heat{4.55} & \textbf{\heat{20.00}} & \textbf{\heat{20.00}} & \heat{26.32} & \textbf{\heat{44.44}} \\
        \bottomrule
    \end{tabular}%
    }
    \label{tab:main_results}
\end{table*}

\section{Video-Browser}
\label{appendix:agent}

\subsection{Experiment Settings}
\label{app:experimental_settings}

\paragraph{Shared setup.}
All offline-track experiments use the 450-question evaluation split and a BM25 index (\(k_1=1.5\), \(b=0.75\), top-10/query) over \{title, description, channel, tags\} on the hard-negative-augmented pool (\nolinkurl{pool_with_hard_negatives}). Transcripts are produced offline by Whisper large-v3-turbo. Accuracy is judged by gpt-5-mini-2025-08-07 (Figure~\ref{fig:eval_prompts}); CE uses $B{=}5$ equally-spaced confidence bins. Each method runs with its defaults below; per-experiment paragraphs list only the parameters that override these defaults.

\paragraph{Video-Browser (ours).}
Planner emits 1 query/loop; Watcher previews each BM25 candidate as $2{\times}3{\times}3$ uniform-frame grids at 256\,px and selects the top-1 in a single multimodal call. 
After each iteration, the Checker evaluates whether the accumulated Watcher-selected evidence is sufficient to answer the question and returns a stopping decision together with feedback for the next search step. If the evidence is insufficient, the feedback is passed to the Planner in the next iteration; otherwise the browsing loop terminates. We cap the loop at $\mathcal{T}_{\max}{=}3$, forcing termination when the maximum number of iterations is reached.
Analyst then runs once at termination over 128 uniformly-sampled frames per accumulated video plus the timestamped transcript (capped at 25k characters). Translation Memory uses the metadata-distilled bank (Appendix~\ref{app:memory_construction}) with BGE-M3 cosine retrieval, top-3 cards from a top-20 proposal pool. Backbones: Qwen3-VL-8B-Instruct (local vLLM) and Gemini-3.1-Pro (API).

\paragraph{ReAct search agents.}
Same search stack, transcripts, and search-and-selection cap as Video-Browser ($\mathcal{T}_{max\_tool\_calling}{=}7$), exposed as a flat ReAct interface with two tools: \texttt{search\_videos} and \texttt{watch}. The \texttt{watch} tool bundles a single dense pass over the chosen video --- 128 uniformly-sampled frames at 1\,fps (long-edge thumbnail 512\,px, JPEG q70) plus the timestamped transcript truncated to 25\,k characters in one multimodal call. 

\paragraph{Adapted video-understanding agents.}
We feed VideoAgent~\citep{wang2024videoagentlongformvideounderstanding}, VideoTree~\citep{wang2025videotreeadaptivetreebasedvideo}, and DVD~\citep{zhang2025deepvideodiscoveryagentic} the top-3 video candidates (\texttt{top\_k\_videos}${=}3$) from the same offline stack, keep their native intra-video loops, and use Qwen3-VL-8B-Instruct throughout. \textit{VideoAgent}: 16 sparse frames per video. \textit{VideoTree}: \texttt{max\_sampled\_frames}${=}64$, branch width 4, up to 4 expansion iterations. \textit{DVD}: 60\,s clips at 10 frames/clip, \texttt{clip\_search\_top\_k}${=}3$, ReAct step budget 3 (16 frames per \texttt{frame\_inspect}, 16 for \texttt{global\_browse}).

\paragraph{Cost decomposition: Video-Browser components.}
Table~\ref{tab:cost_videobrowser} decomposes Video-Browser's per-question token budget into the two phases motivated by its sparse-screening, dense-verification design.
The \textit{Browse} phase --- Planner $+$ Searcher $+$ Watcher $+$ Checker --- represents each BM25 candidate with two $3{\times}3$ uniformly sampled frame grids at 256\,px, and the Watcher selects the top-1 candidate from each returned pool in a single multimodal call.
The browsing loop runs for at most $\mathcal{T}_{\max}{=}3$ iterations.
With Qwen3-VL-8B-Instruct, this scans $18.8$ candidates per question on average for $38.3$\,k tokens, i.e., $\sim 2.0$\,k tokens per candidate browsed. The \textit{Verify} phase is a single Analyst call per question over the accumulated watched set ($3.0$ videos on average), which is fed $128$ uniformly-sampled frames per video plus the timestamped transcript. Verification costs $20.9$\,k tokens per accumulated video on Qwen3-VL-8B-Instruct ($142.5$\,k on Gemini-3.1-Pro). The asymmetry --- $\sim 10\times$ tokens spent per dense-verified video versus per sparsely-browsed candidate --- is the design point of our pipeline: it allows the verification budget to grow with frame coverage without scaling the candidate pool. The Gemini blow-up in absolute terms is driven by Gemini-3.1-Pro's higher per-image tokenisation rather than by additional model calls. Hardware: local Qwen3-VL-8B-Instruct served via vLLM on $4\times$NVIDIA L40s (48\,GB); offline BM25 and Whisper run CPU-side on the same node. Remote backbones (Gemini-3.1-Pro) are accessed through their official APIs.

\begin{table}[h]
    \centering
    \footnotesize
    \setlength{\tabcolsep}{4pt}
    \caption{Per-question token decomposition for Video-Browser, averaged over 450 offline-track questions. ``Browse'' aggregates Planner, Searcher, Stage-1 visual selector, and Checker calls across the $\mathcal{T}_{\max}{=}3$ loops; ``Verify'' is the single Analyst call at termination. ``Tokens / unit'' is per browsed candidate or per verified video. The asymmetry is the intended sparse-vs-dense split of the pipeline.}
    \begin{tabular}{l l c r r r}
        \toprule
        \textbf{Backbone} & \textbf{Phase} & \textbf{Mean per question} & \textbf{Tokens / Q (k)} & \textbf{Tokens / unit (k)} & \textbf{\% of total} \\
        \midrule
        \multirow{3}{*}{Qwen3-VL-8B-Instruct}
            & Browse  & 18.8 candidates browsed & 38.3  & 2.0  &  38\% \\
            & Verify  & 3.0 videos verified     & 62.6  & 20.9 &  62\% \\
            & Total   & ---                     & 100.9 & ---  & 100\% \\
        \midrule
        \multirow{3}{*}{Gemini-3.1-Pro}
            & Browse  & 18.8 candidates browsed &  73.1 &   3.9 &  15\% \\
            & Verify  & 3.0 videos verified     & 427.6 & 142.5 &  85\% \\
            & Total   & ---                     & 500.7 &  ---  & 100\% \\
        \bottomrule
    \end{tabular}
    \vspace{2mm}
    \label{tab:cost_videobrowser}
\end{table}

\subsection{Translation Memory Construction}
\label{app:memory_construction}

Translation Memory is built offline once from the training split before any agent run. Given a training set $D_{\mathrm{train}} = \{(Q_i, A_i, E_i)\}_{i=1}^{N}$, where $Q_i$ is a question and $E_i$ is the metadata of its annotated seed video (title, description, channel, tags, transcript), we want a memory card $c_i = (Q_i, q_i^\star, \alpha_i)$ where $Q_i$ serves as the retrieval key for cosine matching against future user questions, $q_i^\star$ is a BM25 query that has been verified to surface the seed video, and $\alpha_i$ is a small set of anchor entities mined from $E_i$.

\paragraph{Distillation.}
For each row, we render the distillation prompt of Figure~\ref{prompt:memory_distill} and call the LLM once to emit a JSON object with two fields: a list of \emph{exactly five} candidate BM25 queries $\mathcal{C}_i = \{q_{i,1}, \ldots, q_{i,5}\}$, and an anchor block $\alpha_i$ containing 2--4 proper-noun entities, 1--4 salient on-screen / spoken terms, and an optional short channel signal. The prompt enforces hard constraints on each candidate: at most 12 words, no quoting / no Boolean operators, no answer-leak, no numeric timestamps, and no verbatim transcript phrases. The model is asked to silently self-check the constraints before emitting JSON.

\paragraph{BM25 Verification.}
Each candidate $q_{i,j}$ is then issued against the offline BM25 index used at evaluation time (the same \texttt{pool\_with\_hard\_negatives} stack from Appendix~\ref{app:experimental_settings}, indexed over the \textit{title}, \textit{description}, \textit{channel}, and \textit{tags} fields). Let $\mathrm{rank}(q, E_i)$ denote the 1-based rank of the seed video in the top-$K$ BM25 result list for query $q$ ($K{=}3$ in our build), or $\infty$ if it does not appear. The verified query is the candidate with the smallest such rank:
\begin{equation}
    q_i^\star \;=\; \arg\min_{q \in \mathcal{C}_i} \mathrm{rank}(q, E_i),
    \qquad \text{accepted iff } \mathrm{rank}(q_i^\star, E_i) \leq K.
\end{equation}
Candidates whose rank exceeds $K$ are discarded.

\paragraph{Retry on Miss.}
If no candidate from the first distillation pass surfaces the seed video within the top-$K$, the prompt is re-rendered with a non-empty \texttt{feedback} field --- an explicit instruction telling the model that ``prior queries did not surface the target video within the BM25 top-$K$; try more specific proper-noun entities and distinctive on-screen / spoken terms'' --- and the LLM is called a second time. If the retry surfaces the target, the card's status is recorded as \textit{retry\_ok}; otherwise it falls through to the failure mode below.

\paragraph{Anchors-only Failure Cards.}
When even the retry fails to surface the target, we still emit a card with the same retrieval key $Q_i$ but a different payload: the verified-query line is replaced by ``\textit{BM25 verification failed; suggested search terms: $\alpha_i$}''. These cards remain useful at inference time because the anchor entities still expose the topical granularity at which the past task was framed; the Planner is instructed to treat them as hints rather than as direct queries (see Figure~\ref{prompt:memory_block}). The dropped-payload distinction is captured by the card's \texttt{outcome} tag (\texttt{success} for verified cards, \texttt{failure} for anchors-only cards), and the Planner-side memory injection layer surfaces the failure framing literally.

\paragraph{Card Schema.}
Each card is serialized as a \texttt{MemoryCard} with separate \texttt{index\_text} (the verbatim past question $Q_i$, used at planner time for cosine matching) and \texttt{payload\_text} (the verified-query-plus-anchors block of Figure~\ref{prompt:memory_block}). The bank stores the cards as JSONL alongside a precomputed BGE-M3 embedding matrix over the \texttt{index\_text} field, so retrieval at planner time is a single dot-product against the cached matrix.

\paragraph{Bank Statistics.}
The final Translation Memory bank contains 846 cards constructed from the training split. Of these, \textbf{87.4\%} (739) are verified (\textit{ok} or \textit{retry\_ok}) and
\textbf{12.6\%} (107) are anchors-only failure cards.
Among the verified cards, the seed video appears at BM25 rank 1 for 622 queries (84.2\% of verified), rank 2 for 83 (11.2\%), and rank 3 for 34 (4.6\%).
The bank is roughly balanced across the eight Video-BrowseComp genres (Sports / Video \& Shorts / Education \& Knowledge / Film / Documentary /
Games \& Esports / TV Series / Music \& Variety) and across difficulty levels
L1 (476 cards) and L2 (370 cards); L3 questions are excluded from training as they require multi-source evidence chains.

\par\begingroup
\captionsetup{hypcap=false}
\begin{tcolorbox}[colback=gray!10, colframe=gray!50, breakable, title={Memory Distillation Prompt}]
The distillation prompt takes the training question and the seed video's metadata, and asks the LLM to emit five candidate BM25 queries plus anchor terms that can later be verified against the offline search engine. The Jinja placeholders \texttt{\{\{\allowbreak question\allowbreak \}\}}, \texttt{\{\{\allowbreak title\allowbreak \}\}}, \texttt{\{\{\allowbreak channel\allowbreak \}\}}, \texttt{\{\{\allowbreak tags\allowbreak \}\}}, \texttt{\{\{\allowbreak description\allowbreak \}\}}, and \texttt{\{\{\allowbreak transcript\_snippet\allowbreak \}\}} are filled in from each training row; the description is truncated at 1{,}500 characters and the transcript snippet at 800 characters. The optional \texttt{\{\{\allowbreak feedback\allowbreak \}\}} block is only present on the second (retry) call after a verification miss.

\begin{verbatim}
You are designing BM25 search queries that will surface a 
specific YouTube video against a large offline pool. The pool 
is indexed over title, description, channel and tags. 
We will score your queries by whether the target video 
appears in the top results — so concrete entity names 
and salient on-screen / spoken terms matter more than 
generic phrases.

# Target

User question (the kind of question we want to answer using 
this video):
{{ question }}

Seed video metadata (this is the video that contains the 
answer):
- Title: {{ title }}
- Channel: {{ channel }}
- Tags: {{ tags | join(', ') if tags else '(none)' }}
- Description (truncated):
{{ description }}
- Transcript snippet:
{{ transcript_snippet }}

{% if feedback %}# Feedback from previous attempt
{{ feedback }}

{% endif %}# Hard constraints

1. Produce exactly 5 candidate BM25 queries. Each query is a 
   short string (<= 12 words), lowercase preferred, no quoting,
   no operators (no AND/OR/quotes).
2. Diversity: at least two queries should anchor on proper-noun
   entities from title/description; at least one should use
   salient on-screen / spoken terms from the transcript or
   description; the remaining can mix.
3. Do NOT echo any answer-like phrase you can infer from the
   question (e.g., a number, name, or specific event the
   question is asking for), do NOT include numeric timestamps,
   and do NOT include verbatim transcript phrases that
   obviously give away the answer.
4. Anchors block: list 2--4 proper-noun entities, 1--4 salient
   terms (on-screen text, named mechanics, distinctive props),
   and at most one short channel/source signal (or empty
   string if none is distinctive).

# Output

Emit ONLY the following JSON object, no commentary, no code 
fences:

{
  "queries": ["q1", "q2", "q3", "q4", "q5"],
  "anchors": {
    "entities": ["E1", "E2"],
    "channel_signal": "short hint or empty string",
    "salient_terms": ["t1", "t2"]
  }
}

## Self-check (perform silently before emitting)
Before writing the JSON, verify each item:
- [ ] No query contains a numeric timestamp (e.g., "0:24", 
"161s").
- [ ] No query echoes the answer the user is looking for.
- [ ] No query contains a verbatim quoted phrase from the
      transcript.
- [ ] The JSON has exactly 5 entries in "queries".
- [ ] Each query is <= 12 words.
If any check fails, rewrite the affected queries before 
emitting.
\end{verbatim}

\textbf{Retry feedback (rendered into \texttt{\{\{\allowbreak feedback\allowbreak \}\}} on the second attempt only).}
\begin{verbatim}
prior queries did not surface the target video within the BM25
top-K; try more specific proper-noun entities and distinctive
on-screen / spoken terms
\end{verbatim}

\end{tcolorbox}
\captionof{figure}{Memory Distillation Prompt used during offline Translation Memory construction.}
\label{prompt:memory_distill}\par\endgroup

\subsection{Prompts}
\label{agent_prompts}

We provide the prompts used by each LLM-invoked Video-Browser component: Planner (Figure~\ref{prompt:planner}), Translation Memory block injected before the Planner prompt (Figure~\ref{prompt:memory_block}), Watcher (Figure~\ref{prompt:watcher}), Checker (Figure~\ref{prompt:checker}) and Analyst (Figure~\ref{prompt:analyst}).

\par\begingroup
\captionsetup{hypcap=false}

    \begin{tcolorbox}[breakable,colback=gray!10, colframe=gray!50, title={Planner Prompt}]
    The Planner system prompt below is the form that is sent to the LLM whenever Translation Memory retrieval returns at least one card for the current user goal --- the configuration we use throughout the paper. The placeholder \texttt{\{\{ memory\_block \}\}} is filled in at runtime with the rendered memory block (Figure~\ref{prompt:memory_block}); the \texttt{\{\{ max\_queries \}\}} placeholder is filled in with \texttt{config.planner.max\_queries}. The runtime delivers this string as the system message and concatenates the current planner-state view (browsing history, retrieved evidence, prior feedback) as the human message.

\begin{verbatim}
{{ memory_block }}

[How to use Past Experience]
Each entry above was retrieved because its "Past question" is
semantically close to the current User Goal — but the past 
video and the user's current target are DIFFERENT videos. 
The labels mean:
- "Past question": the prior user question this entry was 
  built from. The past target video is not what the current 
  user is looking for.
- "Verified BM25 query": a search string that was returned 
  correct video for that past question. Use it as a query 
  STRUCTURE template —
  its entities belong to the past task; you can learn them with
  entities to solve the current User Goal.
- "Key anchors": entity / proper-noun fragments relevant to 
  past videos.
- "BM25 verification failed; suggested search terms: ...": no 
  working query was confirmed for that past episode; only 
  the entity hints are available.

By studying and reflecting on these past experiences, you can 
figure out how to help users achieve their current goals.

You are the Planner for a video research loop.

Your job is to decide what to search next. Use the user goal, 
search history, text context, and current knowledge state to 
identify the smallest unresolved gaps.

Rules:
1. Do not repeat solved searches.
2. Prefer targeted follow-up queries over broad restarts.
3. If prior evidence surfaced concrete names, entities, dates, 
   or hypotheses, pivot to verifying them directly.
4. Return at most {{ max_queries }} search queries.
5. Output valid JSON only.

Return:
{
  "thought": "brief reasoning about the remaining gap and 
              search strategy",
  "search_queries": ["query1", "query2"]
}
\end{verbatim}

    \end{tcolorbox}
    \captionof{figure}{Planner Prompt}
    \label{prompt:planner}
\par\endgroup

\par\begingroup
\captionsetup{hypcap=false}

    \begin{tcolorbox}[breakable,colback=gray!10, colframe=gray!50, title={Memory Block Injected Prompt}]
    When Translation Memory retrieval returns one or more cards for the current user goal, the runtime renders the following block and prepends it (verbatim) to the Planner's system prompt (Figure~\ref{prompt:planner}). Each card carries (i)~an \textit{index\_text} (the past question, embedded for cosine match against the current user query) and (ii)~a \textit{payload\_text} (a verified BM25 query plus anchor entities mined from the seed video metadata). The injection layer surfaces both: \textit{index\_text} appears as ``Past question'', and \textit{payload\_text} contributes ``Verified BM25 query'' and ``Key anchors''.

\medskip
\textbf{Rendered Memory Block (literal):}
\begin{verbatim}
[Past Experience]
The following lessons were learned from previous similar tasks. 
Use them to guide your strategy, not as hard rules:

1. Past question: "<index_text of card 1>"
Verified BM25 query: <verified BM25 query 1>
Key anchors: <entity 1> · <entity 2> · <salient term> · 
<channel>

2. Past question: "<index_text of card 2>"
Verified BM25 query: <verified BM25 query 2>
Key anchors: <entity 1> · <entity 2>

...

(From N past episodes: K correct, N-K incorrect)
\end{verbatim}

\textbf{Failure-mode card.} When the offline construction pipeline fails to find any BM25 query that retrieves the seed video at acceptable rank, the card is still emitted but with anchor-only payload:
\begin{verbatim}
3. Past question: "<index_text>"
BM25 verification failed; suggested search terms:
   <entity 1> · <entity 2>
\end{verbatim}

\textbf{Concrete example (one card, verified).}
\begin{verbatim}
[Past Experience]
The following lessons were learned from previous similar tasks. 
Use them to guide your strategy, not as hard rules:

1. Past question: "What is the name of the food that 
   protagonist eats in the opening scene of Inception, which 
   symbolizes his relationship with his late wife?"
Verified BM25 query: chasing dreams facing challenges inception 
movie Key anchors: Inception · Christopher Nolan · top 
spinning · movie

(From 1 past episode: 1 correct, 0 incorrect)
\end{verbatim}

The Planner is instructed to treat these cards as structural guidance, not as direct answers: the Verified BM25 query teaches a query \emph{form} that previously surfaced metadata-aligned candidates, while the Key anchors expose the entity granularity at which the past search was framed.

    \end{tcolorbox}
    \captionof{figure}{Memory Block Injected Prompt.}
    \label{prompt:memory_block}
\par\endgroup

\par\begingroup
\captionsetup{hypcap=false}

    \begin{tcolorbox}[breakable,colback=gray!10, colframe=gray!50, title={Watcher Prompt}]
    The Watcher receives the BM25-retrieved candidate pool from the Searcher and selects the top-$k$ candidates whose metadata plus visual previews most plausibly contain the answer. Each candidate is represented by its textual metadata and up to $N$ uniformly-sampled frame grids (default 2 grids of $3{\times}3$ frames at 256\,px). Each cell is annotated with its timestamp.

\medskip
\textbf{Header (single multimodal message):}
\begin{verbatim}
You are screening video search results.
User Query: "{user_query}"

You see {N_candidates} BM25-retrieved candidates. For each, the
metadata is shown below; up to {n_grids} {rows}x{rows} grids
({frames_per_grid} frames each, sampled uniformly across the 
video in temporal order) are attached. Each cell is labelled 
with its timestamp.

Pick the {top_k} most relevant candidate(s) whose metadata + 
visual content most plausibly contain the answer.

Respond with ONLY a JSON list of {top_k} integer candidate 
indices, best first. Example: [3, 0, 7]
\end{verbatim}

\textbf{Per-candidate block (repeated for each candidate $i$):}
\begin{verbatim}
[Cand i] Title: <title> | Channel: <channel>
  Desc: <truncated description>
  [Cand i · Grid 1/N · t=<t_start>-<t_end>s]   <grid 1 image>
  [Cand i · Grid 2/N · t=<t_start>-<t_end>s]   <grid 2 image>
  ...
\end{verbatim}

Title is truncated at 200 characters, description at 240, channel at 60. If a candidate has no decoded grids (download or decode failure), the block falls back to ``\texttt{(no visual data)}'' and the Watcher still scores it on metadata alone.

\medskip
\textbf{Expected output.} A JSON list of \texttt{top\_k} integer indices into the candidate ordering, e.g.\ \texttt{[3, 0, 7]}. The runtime parses the response, deduplicates indices, clips at \texttt{top\_k}, and falls back to the first \texttt{top\_k} candidates if parsing fails or the list is empty.

    \end{tcolorbox}
    \captionof{figure}{Watcher Prompt}
    \label{prompt:watcher}
\par\endgroup

\par\begingroup
\captionsetup{hypcap=false}

    \begin{tcolorbox}[breakable,colback=gray!10, colframe=gray!50, title={Checker Prompt}]
        You are the Checker in an iterative video browsing agent.

Your task is to determine whether the accumulated evidence is sufficient to answer the User Query, or whether another search iteration is needed.

\begin{verbatim}
User Query:
{user_query}

Current Search Query:
{current_query}

Accumulated Evidence:
{accumulated_evidence}

Browsing History:
{history}
\end{verbatim}

If the available evidence is sufficient to support a reliable answer, return HALT. Otherwise, return CONTINUE and briefly identify what evidence is still missing so that the Planner can formulate the next search query.

Return ONLY valid JSON:
\begin{verbatim}
{
  "decision": "HALT" or "CONTINUE",
  "feedback": "brief description of the missing evidence, 
  or empty if HALT"
}
\end{verbatim}

    \end{tcolorbox}
    \captionof{figure}{Checker Prompt}
    \label{prompt:checker}
\par\endgroup

\par\begingroup
\captionsetup{hypcap=false}

    \begin{tcolorbox}[breakable,colback=gray!10, colframe=gray!50, title={Analyst Prompt}]
        After the search loop terminates, the Analyst receives every Watcher-selected video accumulated across all loops. For each video, the runtime extracts \texttt{config.analyst.num\_frames} sparse frames uniformly across the full duration (default 128) and pairs them with the (truncated) timestamped transcript. All videos are concatenated into a single multimodal message; the Analyst is asked to synthesize the final answer in one call.

\medskip
\textbf{Intro (single text block at the head of the message):}
\begin{verbatim}
User Query: {user_query}

Analyze the following video content (transcripts and sampled 
frames) to answer the query.
\end{verbatim}

\textbf{Per-video block (repeated for each video in the verified store):}
\begin{verbatim}
=== Video: <title> ===

[Frame at <t1>s]   <frame image>
[Frame at <t2>s]   <frame image>
...
[Frame at <tK>s]   <frame image>

Transcript:[<t_start>s - <t_end>s] <line>
[<t_start>s - <t_end>s] <line>
...
\end{verbatim}

The transcript is truncated to 25{,}000 characters per video as a safety bound; frame timestamps follow the uniform sample plan and are formatted as ``\texttt{[Frame at \textit{t}s]}'' immediately before each frame image.

\medskip
\textbf{Trailing instruction (single text block appended after all video blocks):}
\begin{verbatim}
Based on the video frames and transcripts provided above, 
answer the User Query.
Response Format:
{
    "Explanation": "your explanation for your final answer",
    "Answer": "your succinct, final answer",
    "Confidence": "your confidence score between 0% and 100% 
                   for your answer"
}
\end{verbatim}

    \end{tcolorbox}
    \captionof{figure}{Analyst Prompt}
    \label{prompt:analyst}
\par\endgroup

\subsection{Failure Case Analysis} 
\label{failure_case}

Despite the significant improvements, Video-Browser still exhibits limitations in specific complex scenarios. We analyze typical failure modes to provide insights for future improvements.

\begin{figure*}[htbp]
  \centering
   \includegraphics[width=1\linewidth]{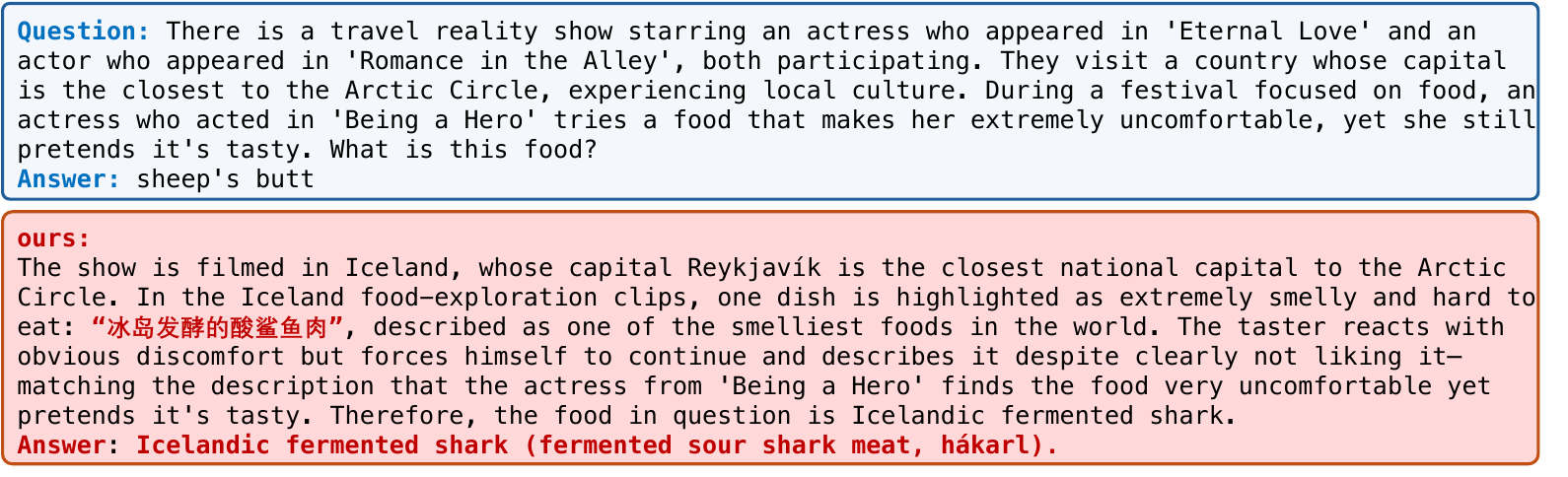}
   
   \caption{Semantic Distractor (Icelandic Food).}
   
   \label{fig:fail3}
\end{figure*}

\noindent\textbf{1. Semantic Distractors (Fig. \ref{fig:fail3}).} In open-ended browsing, multiple videos may share similar semantic attributes. In the Icelandic food case, the agent correctly identified the context (uncomfortable food in Iceland) but was misled by a distractor candidate, fermented shark (hákarl), which shares the attribute of being smelly and hard to eat. The agent failed to distinguish the specific target (``sheep's butt'') from the semantically similar distractor, highlighting the need for stricter entity verification.

\begin{figure*}[htbp]
  \centering
   \includegraphics[width=1\linewidth]{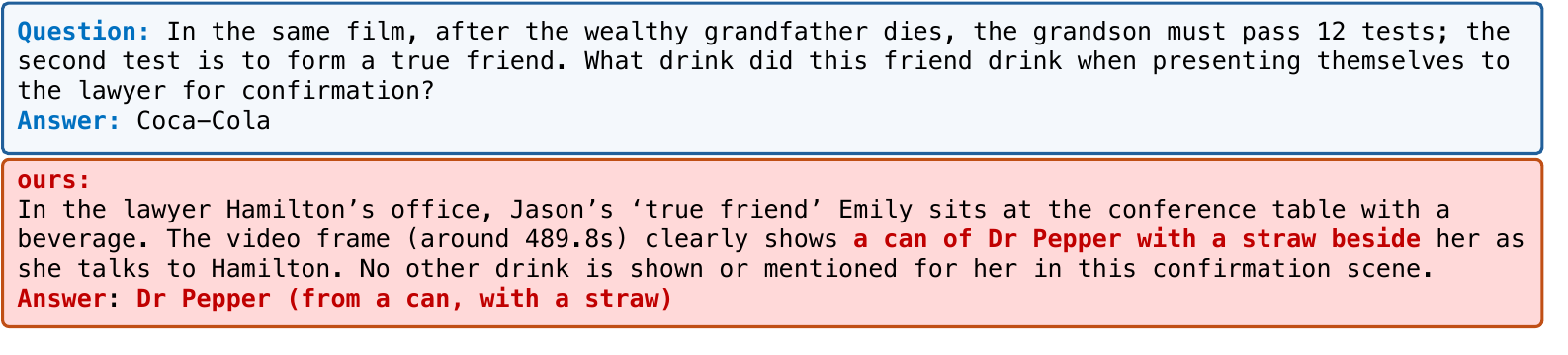}
   
   \caption{Fine-grained Visual Hallucination.}
   
   \label{fig:fail4}
\end{figure*}

\noindent\textbf{2. Fine-grained Visual Hallucination (Fig. \ref{fig:fail4}.)} For extremely small objects, the model may hallucinate specific attributes. In the friendship test case, while the agent correctly located the scene and the object (a beverage can), it misidentified the brand as Dr Pepper instead of the ground truth ``Coca-Cola.'' This suggests that even when the Checker routes the correct video to the Analyst, current MLLMs still struggle with zero-shot OCR or logo recognition on low-resolution objects.

\begin{figure*}[htbp]
  \centering
   \includegraphics[width=1\linewidth]{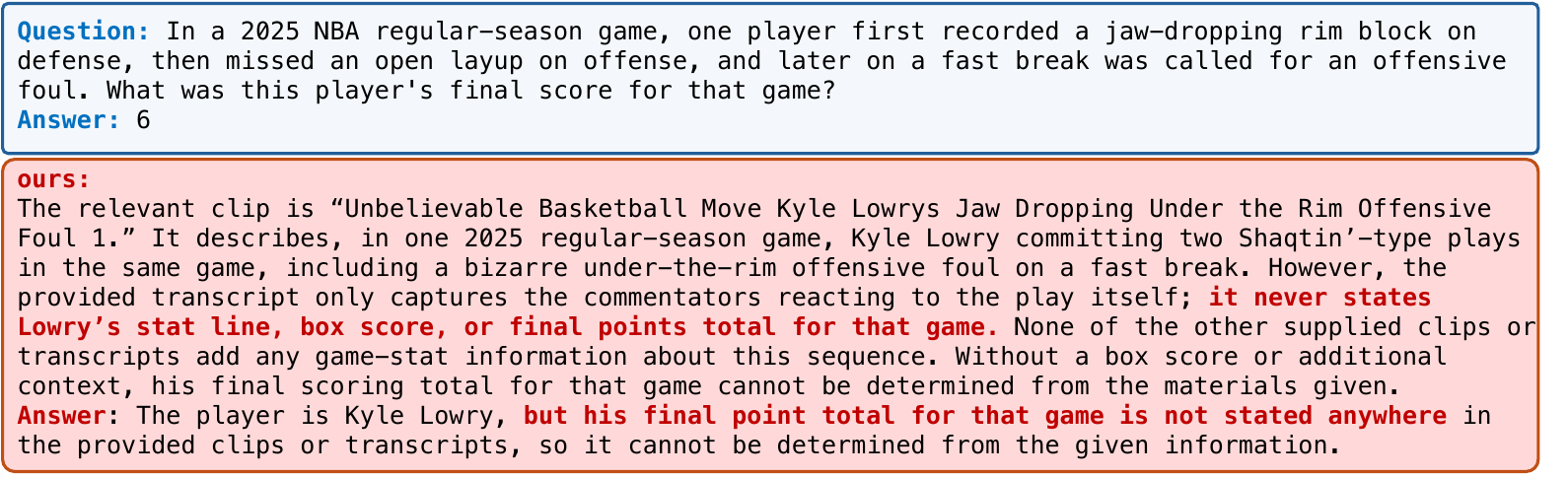}
   
   \caption{Information Deficit (Kyle Lowry).}
   
   \label{fig:fail}
\end{figure*}

\begin{figure*}[htbp]
  \centering
   \includegraphics[width=1\linewidth]{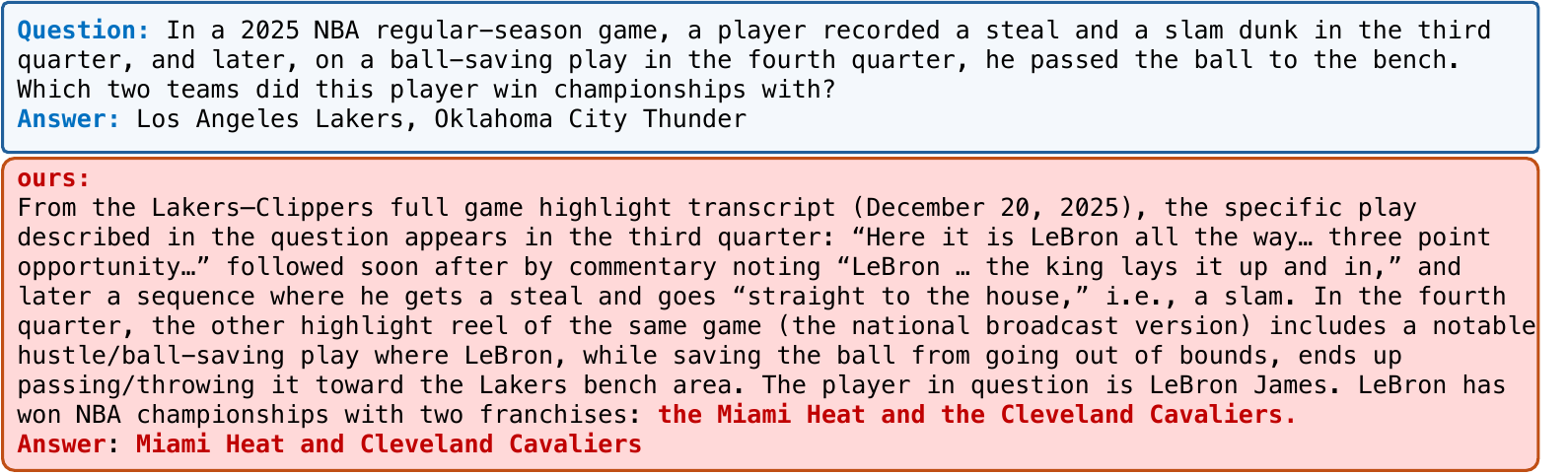}
   
   \caption{Action-to-Entity Mismatch.}
   
   \label{fig:fail2}
\end{figure*}

\noindent\textbf{3. Information Deficit \& Entity Mismatch} The case \ref{fig:fail} (Kyle Lowry) demonstrates an Information Deficit: the agent successfully retrieved the correct highlight clip but failed to answer because the specific statistic (final score) was neither spoken in the transcript nor explicitly displayed in the visual summary. The agent honestly reported not stated, failing to infer or retrieve external box scores. The case \ref{fig:fail2} shows an Action-to-Entity Mismatch. The agent observed the correct basketball plays (block, steal, save) but misattributed them to LeBron James instead of the target player (who played for the Lakers and Thunder), leading to a factually correct answer for the wrong entity.

\clearpage
\section{Limitations and Societal Considerations}
\label{appendix:limitandsocietal}

\subsection{Limitations}
\label{sec:limitations}

Video-BrowseComp and Video-Browser have several limitations. First, the benchmark is built on open-web videos, whose availability, metadata, and platform policies may change over time. To improve reproducibility, our offline track fixes the evaluation corpus and search stack. Second, although Video-BrowseComp covers diverse genres and difficulty levels, it remains a finite benchmark and may not capture all forms of video browsing, especially highly personalized, real-time, or domain-specific search scenarios. Third, the system can still fail when multiple videos share similar metadata and visual cues, when evidence is distributed across weakly linked sources, or when fine-grained visual details are missed by the sparse Watcher.

\subsection{Broader Impacts}
\label{sec:broader_impacts}

Video-BrowseComp and Video-Browser are intended to support reproducible research on evidence-grounded video browsing agents. Potential positive impacts include improving the transparency, reliability, and comparability of systems that search for and verify information from open-web videos. By providing a fixed offline search stack, annotations, prompts, and evaluation protocols, the benchmark can help researchers diagnose whether failures arise from retrieval, candidate selection, or final video reasoning.
At the same time, stronger video-browsing agents may introduce negative societal risks. Because the task involves searching and inspecting open-web videos, such systems could be misused for large-scale information extraction, privacy-invasive profiling, surveillance-like applications, or rapid discovery of sensitive public content. Incorrect retrieval or visual verification may also lead to misidentification, unsupported conclusions, or over-reliance on inaccurate video-grounded answers. These risks are especially relevant when evidence involves people, events, or fine-grained visual attributes that models may misread.
We mitigate these risks by focusing our release on a controlled offline benchmark, documenting data provenance and limitations, avoiding redistribution of third-party raw videos unless permitted by the original licenses, and releasing source identifiers, metadata manifests, reconstruction scripts, prompts, and evaluation code for reproducible research use.

\subsection{Responsible Release, Safeguards, and Asset Documentation}
\label{sec:responsible_release}

\paragraph{Safeguards.}
We do not release a new pretrained generative model. The released assets are a benchmark, a fixed offline search stack, prompts, evaluation code, and metadata needed for reproducible agentic video browsing research. Since the benchmark is built from open-web videos, we avoid repackaging or redistributing raw third-party videos unless redistribution is permitted by the original licenses or terms of use. For videos that cannot be redistributed, we provide source identifiers, metadata manifests, reconstruction scripts, and verification metadata rather than raw video files.

\paragraph{Existing assets and licenses.}
The benchmark uses existing open-web videos, pretrained language and multimodal models, retrievers, and video-agent baselines. We cite the corresponding model, retriever, and baseline papers in the main text and references. The release package will include a license and provenance table specifying, for each asset type, its source, intended use, license or terms-of-use information when available, and whether the asset is redistributed or only referenced through source identifiers. Third-party videos remain governed by their original platform terms and copyright restrictions.

\paragraph{New assets.}
The new assets introduced by this work include Video-BrowseComp annotations, train/evaluation splits, evidence chains, fixed offline search-stack metadata, hard-negative metadata, Translation Memory construction scripts, prompts, and evaluation code. We document the construction pipeline, annotation criteria, quality-control procedure, dataset statistics, evaluation metrics, experimental settings, prompts, and known limitations in the paper and appendix. The released package will include a README, data schema, reconstruction instructions, and license information.

\end{document}